\documentclass{article}

\usepackage[accepted]{icml2025}

\usepackage{myaistats}
\usepackage{mynotation}
\usepackage{local-notation}
\usepackage{xcolor}
\usepackage{enumerate}

\newcommand{\myparagraph}[1]{\paragraph{#1}}

\begin{document}

\twocolumn[
\icmltitle{On the Impact of Performative Risk Minimization for
Binary Random Variables}



\icmlsetsymbol{equal}{*}

\begin{icmlauthorlist}
\icmlauthor{Nikita Tsoy}{insait}
\icmlauthor{Ivan Kirev}{insait}
\icmlauthor{Negin Rahimi}{insait}
\icmlauthor{Nikola Konstantinov}{insait}
\end{icmlauthorlist}

\icmlaffiliation{insait}{INSAIT, Sofia University ``St. Kliment Ohridski''}

\icmlcorrespondingauthor{Nikita Tsoy}{nikita.tsoy@insait.ai}

\icmlkeywords{Performative prediction, impact, sequential decision making}

\vskip 0.3in
]



\printAffiliationsAndNotice{}  

\begin{abstract}
    Performativity, the phenomenon where outcomes are influenced by
    predictions, is particularly prevalent in social contexts where individuals
    strategically respond to a deployed model. In order to preserve the high
    accuracy of machine learning models under distribution shifts caused by
    performativity, \citet{p20p} introduced the concept of performative risk
    minimization (PRM). While this framework ensures model accuracy, it
    overlooks the impact of the PRM on the underlying distributions and the predictions of the model.
     In this paper, we
    initiate the analysis of the impact of PRM, by studying
    performativity for a sequential performative risk minimization problem with binary random variables and linear performative shifts. We formulate two natural measures of impact. In the case of full information, where the distribution
    dynamics are known, we derive explicit formulas for the PRM solution and
    our impact measures. In the case of partial information, we provide
    performative-aware statistical estimators, as well as simulations. Our
    analysis contrasts PRM to alternatives that do not model data shift and
    indicates that PRM can have amplified side effects compared to such methods.
\end{abstract}

\section{Introduction}
\label{sec:intro}

Predictions can significantly influence everyday life \citep{r18t}, an effect
known as performativity. For instance, traffic predictions can alter people's
daily routes, crime predictions can affect police resource allocation, and
stock price predictions can steer traders' decisions. These changes can lead to
shifts in the underlying data distribution, making the original predictions
less accurate.

To capture these effects, \citet{p20p} introduced the concept of
\emph{performative prediction}. In this framework, a deployed model $\prm$
induces a data distribution $D(\prm)$, which gives rise to a new learning
objective, the \emph{performative risk} $\E_{z \sim D(\prm)}(\ell(\prm, z))$,
with $\ell$ being a loss function and $z$ being a sample from the model-induced
distribution. Much progress has been achieved in performative risk
minimization (PRM), i.e., finding performatively-optimal points \citep[see][for a recent survey]{hardt2023performative}.

While PRM is preferable to standard risk minimization (RM) for the sake of test-time accuracy, the broader impact of PRM remain elusive. In particular, using PRM instead of RM leads to different predictions deployed by the learner and also changes the evolution of the data distribution, and these effects are compounded when deploying multiple models over time. This limited understanding of the impact of PRM on the predictions and distribution may be partially due to the mathematical challenges arizing from analyzing the long-term dynamics of the learned models, in the presense of intricate dependencies of the data distribution on all previous models.

\myparagraph{Contributions} In this work, we initiate the analysis of the broader impact of PRM, by studying a sequential performative mean estimation problem for binary variables, in the presence of linear performative distribution shifts. The simplicity of the learning setup enables us to derive the long-term dynamics of PRM, despite the complicated downstream impact of each deployed prediction on the future data distributions. This in turn allows us to quantify the evolution of the predictions and the data distribution.

Within this model, we formulate two measures of impact. The first measure concerns the model predictions and corresponds to the usual statistical notion of a bias of an estimator. The second measure quantifies the shift in the mean of the binary random variable, relative to the mean in the case of lack of performative effects, and thus allows us to understand the evolution of the data distribution under PRM. 

We analyze PRM and the two measures in a one-period (single model deployment) and an infinite horizon (sequential model deployment) setting. In each case, we first study a full
information setting, where all problem parameters (e.g. strength of performativity and initial distribution) are known to the model provider, in
order to isolate the effects of performativity from exploration. We then analyze performativity and exploration jointly via theory and simulations.

Our results indicate that, compared to RM, PRM may select more biased estimators and/or ones that shift the mean to extreme values. This happens in particular because minimizing the PRM loss suggests trading-off the usual mean squared error (MSE) for reduced aleatoric uncertainty in the future data distribution. Such effects occur when the distribution responds positively to model
predictions or when the distribution responds negatively, but the model is
updated rapidly and the performativity is high.

Finally, we use two example scenarios to interpret our measures and technical
results in a social context.\footnote{Please see the replication files for our
paper at
\url{https://github.com/insait-institute/performative-prediction-impact-replication}}

\section{Related Work}

\myparagraph{Performative Prediction}
In machine learning, performativity is often studied within the framework of
\emph{performative prediction}, where the goal is to find a model with good
performance on the distribution that it induces. The setting was introduced by
\citet{p20p} and was inspired by works on strategic classification
\citep{h16s,d04a}. Numerous works study methods for finding performatively
optimal/stable models \citep{mendler2020stochastic,m21o,j22r,
izzo2022learn,ray2022decision,lin2024plugin}, see \citet{hardt2023performative}
for a recent overview. \citet{b22p,r22d,m23p} extend this framework to stateful
environments, where previous model deployments impact the data distribution at
later stages.

In contrast to the works above, we focus on the impact of PRM on the data
distribution and on the predictions made by these models. To our awareness, the
only work that studies properties beyond performative loss in the context of
performative prediction is that of \citet{jin2024addressing}, who, however,
focus on the fairness and polarization properties of PRM instead.

\myparagraph{Distribution Steering}
Our work analyzes the secondary effects of PRM on the distribution and
outcomes, which were not intended by the model provider. Several related works
could allow the model provider to encode penalties for these unintended changes
into its optimization task. \citet{k23m} investigate how to steer distributions
towards a more desirable outcome by using omnipredictors. Similarly,
\citet{g24o} study the task of distribution steering in population dynamics
context. However, these results do not inform about the unintended distribution
changes due to PRM, which is the focus of our work.

\myparagraph{Long-Term Fairness} The line of works on long-term fairness also studies the evolution of
distribution in social contexts. \citet{ensign2018runaway,bechavod2019equal}
focus on social feedback loops. \citet{williams2019dynamic, liu2020disparate}
propose models for performative responses motivated by their learning context.
While these works model performativity, they focus on finding fair models. In
contrast, we focus on performatively optimal algorithms and analyze their
impact on the data distribution and predictions. 

\myparagraph{Instances of Performativity} Performativity arises in many social
contexts. Economic agents respond to the actions of the
government \citep{l76e}. Performative policing affects the distribution of
observed crime rates \citep{e18r}. Traffic predictions reroute drivers to new
areas \citep{m19w,c22i}. Recommendation systems affect the consumption of new
content \citep{b22d,d22p}. Since performativity is so widespread, it is
important to study optimization formulations in such settings and the effects
of performatively-optimal solutions on their environment.

\section{Model}

We now discuss the sequential performative prediction framework we study, the specific instance considered in our analysis and the impact metrics we focus on.

\subsection{Performative Prediction Framework}
\label{subsec:perf_pred_framework}

\myparagraph{Optimization Problem} This work analyzes how optimizing for performative accuracy influences the model provider actions and
the underlying probability distribution. To answer this question, we consider
the following sequential performative prediction problem inspired by
\citet{p20p,b22p,r22d}. Denote by $\Prm$ the space of models, by $\prm_t \in
\Prm$ the model parameters at time $t$, by $\mathcal{D}$ the space of all data
distributions on a data space $\mathcal{Z}$, by $D_t \in \mathcal{D}$ the data
distribution after the response to $\prm_{t-1}$, and by $\timeop\colon
\mathcal{D} \times \Prm \to \mathcal{D}$ the model of performative response,
such that $D_t = \timeop(D_{t-1}, \prm_{t-1})$. The model provider is
interested in minimizing a discounted loss
\begin{equation}
    \label{eq:opt-cont-prob}
    \min_{(\prm_t)_{t=0}^{T-1}} \E_{(\prm_t)_{t=0}^{T-1}}\Par*{\sum_{t=0}^{T-1}
    \gamma^t \E_{\z \sim D_t^\text{test}} (\ell(\prm_t, \z))},
\end{equation}
where $\ell$ is the loss function, $\gamma \in (0, 1)$, $T \in \mathbb{N} \cup
\{\infty\}$, and $\prm_t$ depends only on the information up to time $t$.
We denote the solution to this problem by $(\prm^*_t)_{t=0}^{T-1}$ and refer to it as the \emph{PRM path}. Conceptually, this path can be seen as the ``performatively-optimal'' sequence of models \citep{p20p}.

\begin{remark}
    Notice that our problem falls within the framework of
    reinforcement learning under partial observability, where $\prm_t$
    corresponds to action and $D_t$ corresponds to state.
\end{remark}

\myparagraph{Test Distribution}
The objective (\ref{eq:opt-cont-prob}) depends on the model of test
distributions $D_t^\text{test}$. In the standard performative setting \citep{p20p},
$D_t^\text{test} = D_{t+1}$, the model is tested in an environment adapted to
it. This property holds when the speed of model deployment is slower than that
of societal adaptation. Thus, we call this case the \emph{slow deployment}
case. For example, drug efficacy estimates can only be updated after
time-consuming clinical trials.

We also consider the case of $D_t^\text{test} = D_t$, when the environment
adapts to the predictions with delay. Such delays arise whenever models
are updated frequently. Therefore, we call this case the \emph{rapid
deployment} case. For example, the predictions of road congestion can be
updated ``on the fly'', so people may not be able to adapt to the latest
predictions.

\subsection{Instance of Performative Problem}
\label{subsec:perf_shift_model}

\myparagraph{Distribution} We assume that $D_t$ describes binary random variables $z \sim D_t$ with mean
$p_t$
\[
    z =
    \begin{cases}
        -1/2, & \text{w.p.} \: 1/2-p_t,\\
        1/2, & \text{w.p.} \: 1/2+p_t.
    \end{cases}
\]
Note that $z$ is a Bernoulli random variable shifted by $1/2$ for mathematical
convenience. For these variables, a positive outcome could mean that a drug is
effective for treating a patient or that a certain route is free from traffic.

\myparagraph{Loss} At time $t$, the model provider deploys $\prm_t \in [-\nicefrac{1}{2},
\nicefrac{1}{2}]$ to minimize mean squared error (MSE) $\ell(\prm_t, z) \defeq
(\prm_t - z)^2$. We denote the expected loss (w.r.t. all randomness) by
\[
    \loss_t \defeq \E(\E_{z \sim D^\text{test}_t}((\prm_t - z)^2)).
\]
We denote the means produced by the PRM path $(\prm^*_t)_{t=0}^{T}$ by $(p^*_t)_{t=0}^T$. We also denote by $p^\text{test}_t$ the mean of the distribution $D^\text{test}_t$. Note that $p^\text{test}_t$ is equal to $p_{t+1}$ and $p_t$ in the slow and
rapid deployment cases, respectively.

\begin{lemma}[Error-Uncertainty Tradeoff]
    \label{lemma: bias-variance}
    The mean squared error of $\prm_t$ on $D^{\text{test}}_t$ is
    \begin{equation}
        \label{eq:mse}
        \E\Par{(\prm_t - z)^2 \given \prm_t, p^\text{test}_t} = (\prm_t -
        p^\text{test}_t)^2 + (\nicefrac{1}{4} - (p^\text{test}_t)^2).
    \end{equation}
\end{lemma}

Here, the first term corresponds to the standard mean squared error (MSE).
The second term corresponds to the aleatoric uncertainty of $D^\text{test}_t$
\citep[note that such decompositions are valid for a big class of
distributions,][]{g22e}. Thus, under performativity, the model provider is also incentivized to decrease the environment uncertainty, while in the
non-performative case they only minimize the MSE.

\myparagraph{Performative Response}
Performativity manifests differently in different contexts. For example,
route congestion estimates might have \emph{negative feedback} on the
congestion: when the model predicts that one route is less
busy than others, people might use it more. On the
other hand, drug efficiency estimates might have
\emph{positive feedback} on the drug efficacy due to the well-known placebo
effect.

We capture these effects using a linear response model
\begin{equation}
\label{eqn:linear_response}
    p_{t+1} \defeq \alpha \prm_t + (1 - \abs{\alpha}) s_{t+1},
\end{equation}
where $s_{t+1} \defeq \lambda p_t + (1 - \lambda) \pi$, $\alpha \in (-1, 1)$,
$\lambda \in [0, 1)$, and $\pi \in [-1/2, 1/2]$. Here, $s_{t+1}$ is the next
period mean in the absence of performativity, $\alpha$ controls the strength
and direction of performativity, $\lambda$ controls the friction in the
distribution update, and $\pi$ is the equilibrium (long-term) mean in the
absence of model influence. Positive $\alpha$ describes positive feedback
situations. Negative $\alpha$ describes negative feedback situations. We also
use the notation $\beta \defeq (1 - \abs{\alpha}) \lambda$, under which $p_{t+1} = \alpha \prm_t + \beta p_t + (1 - \abs{\alpha} - \beta) \pi$.

\paragraph{Limitations} This work considers a specific instance of our general performative
framework to get a comprehensive theoretical description of the considered
impact metrics (see \cref{subsec:measures_of_impact}). While the considered form of the problem limits generality, we
believe that our analysis could be informative for real-world situations. Therefore, we
discuss the potential applications and limitations of our analysis in detail in
\cref{sec:conclusion}.

\subsection{Measuring the impact of PRM}
\label{subsec:measures_of_impact}

The distribution we consider is determined by its mean. This property allows
us to formulate two natural ``impact'' metrics, \textit{bias} and
\textit{mean shift}. Bias captures the \textit{impact of PRM on the
learner's predictions}, while mean shift describes the \textit{impact of
PRM on the underlying distribution}. We discuss the generalizations of these
metrics to other learning tasks in \cref{sec:gen-metrics}.

\myparagraph{Bias} Consider an arbitrary path (sequence of predictions) $(\theta_i)_{i=0}^T$. Inspired by the classic notion of bias, at each time $t$ we study the expected error in the
estimate of the mean
\begin{align}
\label{eqn:bias}
    \bias_t \defeq \E\Par{\prm_t - p^\text{test}_t}.
\end{align}
Intuitively, the bias captures how far (on average) are the predictions of the path from the true mean at a given time.

\myparagraph{Mean Shift} Here we compare the mean $p_t$ of the distribution under the path $(\theta_i)_{i=0}^T$ and corresponding mean in the absence of
performativity $p^0_t$ (i.e., when $\alpha = 0$). Formally,
\begin{equation}
    \label{eqn:shift}
    \shift_t \defeq \E\Par{p_t - p^0_t}.
\end{equation}
The mean shift measures the amount (and direction) of deviation of the mean of the distribution under the considered path $(p_i)_{i=0}^T$, compared to mean $p^0_t$ at time $t$ if the distribution was not affected by the predictions.

\myparagraph{Analyzing the impact of PRM}
We study the bias and shift of the PRM path $(\prm^*_t)_{t=0}^{T-1}$, which we
denote by $\bias^*_t$ and $\shift^*_t$, respectively. Additionally, we
compare the PRM path to a \emph{naive} path, $\prm^n_t$, which ignores the
performativity when making predictions. Formally, $\prm^n_t$ is defined as the
mean of the previously observable distribution
\begin{equation}
    \label{eqn:naive_path}
    \prm^n_t \defeq p^\text{test}_{t-1}.
\end{equation}
This corresponds to the usual approach to prediction in which one minimizes the
loss with respect to the currently observable distribution (akin to the usual
ERM principle). In the rapid case, where no distribution is observed in the
first period, we define $\prm^n_0 = 0$. We will denote the bias and
shift of the naive path by $\bias^n_t$ and $\shift^n_t$, respectively.

\myparagraph{Interpreting bias and shift}
Sections \ref{sec:one_period} and \ref{sec:infinite_horizon} derive explicit
formulas for our impact measures under the PRM and naive paths, allowing us to
reason about the quantitative behaviour of these metrics. In general, high bias
can be interpreted as an undesirable property, even from a solely statistical
standpoint \cite{young2005essentials}. However, as we will see, the PRM path is
biased due to the trade-off with distribution uncertainty (Lemma \ref{lemma:
bias-variance}). In contrast, mean shift interpretation is usually
context-dependent. \cref{sec:conclusion} provides two example scenarios to
illustrate these points.

\section{One-Period Model}
\label{sec:one_period}

This section analyzes the case of $T=1$. First, we discuss the full information
case where $p_0$, $\alpha$, $\lambda$, and $\pi$ are known to the model
provider. This allows us to separate the effects of PRM
from the hardness of designing of algorithms that achieve PRM (due to exploration/finite-sample effects). Next, we assume that the initial mean $p_0$ is unknown and study how this
uncertainty affects our previous results. Finally, in an episodic RL
setting, we study via simulations the case where no information about the
parameters is available. 

Notice that the slow deployment case for $T=1$, which
is the main focus of this section, corresponds to the standard setting of
\cite{p20p}.

\subsection{Perfect Information}

\subsubsection{Slow Deployment}
\label{sec:one-slow}

\begin{proposition}[Proof in \cref{sec:proof-one-slow-sol}]
    \label{thm:one-slow-sol}
    The solution to the problem (\ref{eq:opt-cont-prob}) in the $T=1$ slow deployment case is
    \[
        \prm^*_0 =
        \begin{cases}
            \clip\Par[\big]{\frac{(1 - \abs{\alpha}) s_1}{1 - 2 \alpha},
            -\frac{1}{2}, \frac{1}{2}}, & 1 - 2 \alpha > 0,\\
            \sign(s_1) / 2, & 1 - 2 \alpha \le 0.
        \end{cases}
    \]
    Whenever $\abs{\prm^*_0} \neq 1 / 2$, we get
    \[
        p^*_1 = \frac{(1 - \alpha) (1 - \abs{\alpha})}{1 - 2 \alpha} s_1 =
        \frac{1 - \alpha}{1 - 2 \alpha} (\beta p_0 + (1 - \abs{\alpha} - \beta)
        \pi).
    \]
\end{proposition}

We visualize this solution in \cref{fig:one-slow-sol}. For the rest of the subsection we assume that $\prm^*_0 \neq 1/2$.

\begin{figure}[ht]
    \centering
    \input{fig1-one-slow.pgf}
    \caption{The dependence of $\prm^*_0$ (blue), $p^*_1$ (orange), and $s_1$
    (green) on $p_0$ for $\lambda = 0.8$ and $\pi = 0.2$ in slow $T=1$ case.
    Columns correspond to the different $\alpha$.}
    \label{fig:one-slow-sol}
\end{figure}

\myparagraph{Loss} We get that
\[
    \loss^*_0 =
    \begin{cases}
        \frac{1}{4} - \frac{(1 - \abs{\alpha})^2 s_1^2}{1 - 2 \alpha},
        & \abs{s_1} < \frac{\nicefrac{1}{2} - \alpha}{1 - \abs{\alpha}},\\
        \frac{1 - \alpha}{2} - (1 - \abs{\alpha}) \abs{s_1},
        & \abs{s_1} \ge \frac{\nicefrac{1}{2} - \alpha}{1 - \abs{\alpha}}.
    \end{cases}
\]

\myparagraph{Bias} We have that $\bias^*_0 = \frac{\alpha (1 - \abs{\alpha}) s_1}{1 - 2 \alpha}.$
Thus, the PRM path is generally biased. This bias does not arise from the usual
bias-variance trade-off in statistics. Instead, the performativity incentivizes
the model provider to reduce the uncertainty in the distribution. To see this, notice that the unbiased predictor, which minimizes the error term in MSE
(\ref{eq:mse}), exists: $\prm^u_0 = \frac{1 - \abs{\alpha}}{1 - \alpha} s_1$.
For positive feedback, the prediction is biased towards extreme values
(i.e., $-1/2$ or $1/2$). For negative, the prediction is biased towards
$0$. The absolute value of bias increases in $\abs{\alpha}$ if $\alpha >
-\frac{\sqrt{3} - 1}{2}$.

\myparagraph{Shift}
Once $\prm^*_0$ is deployed, it induces shift $\shift^*_1 = \frac{\alpha -
\abs{\alpha} + \alpha \abs{\alpha}}{1 - 2\alpha} s_1$.
The direction of the shift depends on $\sign(\alpha)$ in the same way as the
bias. The effect increases with $\abs{\alpha}$. While a no-shift prediction
exists, $\prm_0 = \sign(\alpha) s_1$, it differs from the unbiased prediction
in the negative feedback case. This shows that unbiasedness and the
absence of shift can not be achieved simultaneously under negative feedback.

\myparagraph{Discussion} We can see that, in general, the PRM prediction is biased (even though the
model provider has perfect information about the distribution), and its impact
on the mean of the distribution is not zero. In the positive feedback case, the
model provider benefits from shifting the mean to extreme values. Even though
this strategy increases the error term, it decreases the uncertainty. In the
negative feedback case, the performative response to the unbiased prediction
shifts the mean closer to $0$. So, the provider employs a biased prediction to
reduce this drop in the uncertainty.

\myparagraph{Comparison with Naive Path} Now, we consider the naive path, where, for the clarity of exposition, we
assume that the system was initially at equilibrium, i.e., $p_0 = \pi =
s_1$. We get
\[
    \begin{split}
        p^n_1 &= (1 + \alpha - \abs{\alpha}) s_1,\\
        \loss^n_0 &= 1/4 - (1 + 2 \alpha - 2 \abs{\alpha}) s_1^2,\\
        \shift^n_1 &= -\bias^n_1 = (\alpha - \abs{\alpha}) s_1.
    \end{split}
\]
If $\alpha > 0$, the bias and shift of the naive path is zero, which might be
more desirable compared to the PRM path. However, the naive loss is worse
than the PRM loss by $\frac{\alpha^2}{1 - 2 \alpha} s_1^2$.  At the same
time, if $\alpha \le 0$, the bias and shift of the naive path are higher in
absolute values than the bias and shift of the PRM path, i.e. RM increases our measures in the negative feedback case compared to PRM.
Moreover, in the negative case, the loss penalty increases to $\frac{9
\alpha^2}{1 - 2 \alpha} s_1^2$.

\begin{figure*}[ht]
    \includegraphics[width=\linewidth]{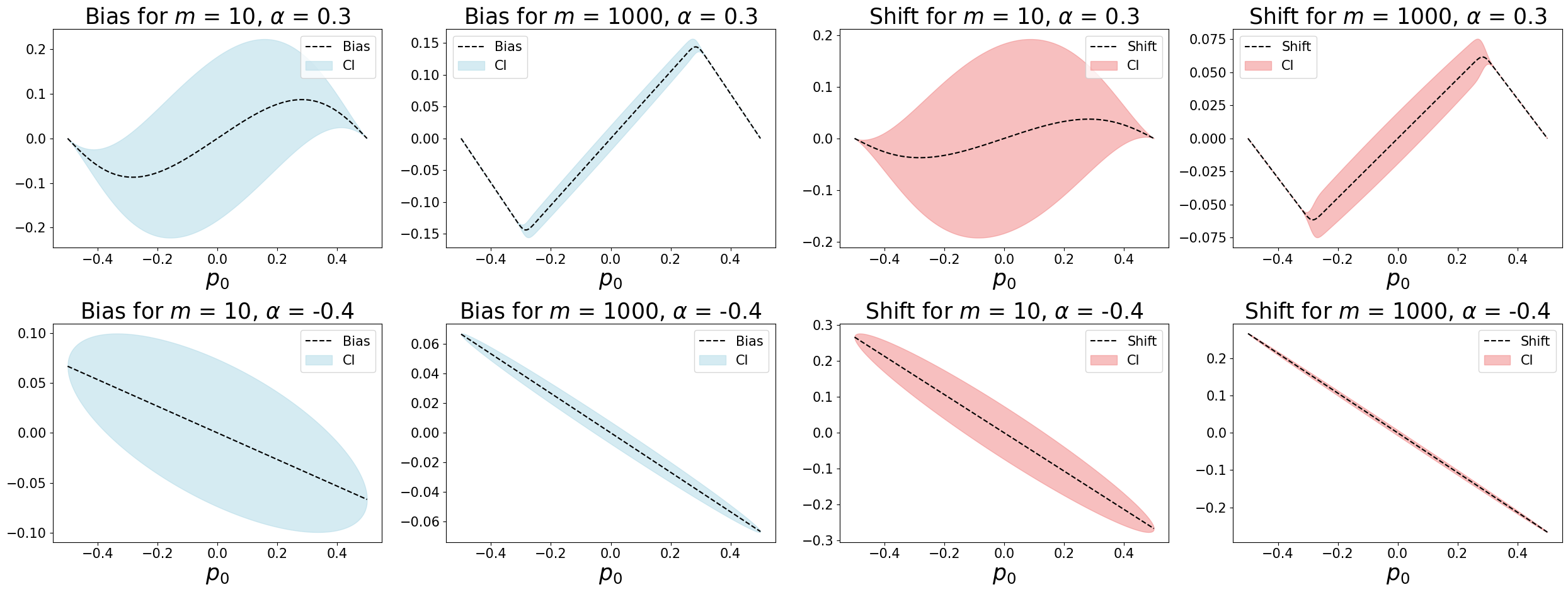}
    \caption{The dependence of $\bias(\hat{\prm}^*_0)$ (left) and
    $\shift(\hat{\prm}^*_0)$ (right) and corresponding variances on $p_0$. The
    upper row corresponds to $\alpha=0.3$, the lower row corresponds to
    $\alpha=-0.4$. Columns correspond to the different $m$.}
    \label{fig: bias_shift}
\end{figure*}

\subsubsection{Rapid Deployment}

\myparagraph{One-Period Case} \cref{eq:opt-cont-prob} reduces to
\[
    \min_{\prm_0 \in [-\nicefrac{1}{2}, \nicefrac{1}{2}]} \prm_0^2 - 2 \prm_0
    p_0,
\]
which results in $\prm^*_0 = p_0$. (We visualize this solution in
\cref{fig:fin-sols}, top row, in Appendix.) Additionally, we get
\[
    \begin{split}
        p^*_1 &= (\alpha + \beta) p_0 + (1 - \abs{\alpha} - \beta) \pi,\\
        \bias^*_0 &= 0,\\
        \shift^*_1 &= (\alpha - \abs{\alpha} \lambda) p_0 - \abs{\alpha} (1 -
        \lambda) \pi.
    \end{split}
\]
If $\alpha > 0$, the PRM prediction shifts the mean closer to $p_0$
relative to $\pi$. If $\alpha < 0$, the effect is hard to interpret. We
only consider the case of $\pi = p_0$. In this case, $\sign(p_0) \neq
\sign(p^*_1 - s_1)$. The mean is shifted away from $p_0$ in the direction of
$0$. Also note that the absolute value of the shift increases in
$\abs{\alpha}$ under both negative and positive feedback.

\myparagraph{Comparison with Two-Period Case} To see whether the prediction remains unbiased once the distribution changes, we
compare the one- and two-period models. For $T=2$, we get the following two-period
problem:
\[
    \min_{\prm_0, \prm_1, p_1 \in [-\nicefrac{1}{2}, \nicefrac{1}{2}]}
    \sum_{t=0}^1 \gamma^t (\prm_t^2 - 2 \prm_t p_t)
\]
such that $p_1 = \alpha \prm_0 + \beta p_0 + (1 - \abs{\alpha}) (1 - \lambda)
\pi$.

\begin{proposition}[Proof in \cref{sec:proof-two-rapid-sol}]
    \label{thm:two-rapid-sol}
    The solution to the problem (\ref{eq:opt-cont-prob}) in the $T=2$ rapid deployment case is
    \[
        \begin{split}
            \prm^*_0 &= \clip\Par[\Big]{\frac{(1 + \gamma \alpha \beta) p_0 +
            \gamma \alpha (1 - \abs{\alpha} - \beta) \pi}{1 - \gamma \alpha^2},
            -\frac{1}{2}, \frac{1}{2}},\\
            \prm^*_1 &= p^*_1.
        \end{split}
    \]
    Whenever $\abs{\prm^*_0} \neq 1 / 2$, we get
    \[
        p^*_1 = \frac{(\alpha + \beta) p_0 + (1 - \abs{\alpha} - \beta) \pi}{1
        - \gamma \alpha^2}.
    \]
\end{proposition}

\cref{fig:fin-sols}, middle row, in Appendix visualizes $\prm^*_0$. If $\abs{\prm^*_0} < 1/2$, we get
\[
    \begin{split}
        \bias^*_0 & = \frac{\gamma \alpha (\alpha + \beta)
        p_0 + \gamma \alpha (1 - \abs{\alpha} - \beta) \pi}{1 - \gamma
        \alpha^2},\\
        \shift^*_1 & = \frac{(\alpha \! - \! \abs{\alpha} \lambda \! + \!
        \gamma \alpha^2 \lambda) p_0 - (\abs{\alpha} \! - \! \gamma \alpha^2)
        (1 \! - \! \lambda) \pi}{1 - \gamma \alpha^2},\\
        \bias^*_1 & = 0.
    \end{split}
\]
Compared to the case of $T=1$, the mean shifts to more extreme values due
to the denominator, and the first-period bias becomes non-zero. However, the
final prediction remains unbiased. The long-term loss incentivizes the
model provider to actively manipulate the mean, even if the short-term loss
suffers from such manipulation.

\myparagraph{Summary}
Similarly to the slow case, the bias and shift of the PRM path are
generally not zero and increase in $\abs{\alpha}$. In contrast to the
slow case, only the long-term effects incentivize uncertainty optimization in
the rapid model.

\subsection{Imperfect Information}
\label{sec:imperf-one-slow}

This section analyzes how uncertainty affects our full information results in
the slow deployment case. 

\subsubsection{Unknown Mean}

First, we analyze the case where $\alpha$ and $\lambda$ are known to the
model provider but $p_0$ is unknown. Thus, the model provider needs to
simultaneously learn $p_0$ and adjust for performativity. For simplicity, we
focus on the equilibrium case where $p_0 = \pi$. To learn $p_0$, the model
provider observes $m$ i.i.d. samples $S_0 = \{p_{0, i} \}_{i=1}^m \sim D_0^m$
and uses an estimator $\theta_0\colon \mathbb{R}^m \to [-1/2, 1/2]$ to
get an estimate $\theta_0(S_0)$.

\myparagraph{Estimators} To study the extent to which the results of the previous section transfer, we
introduce the analogues of the optimal and naive predictions. For the naive
case, we use the empirical mean $\hat{\theta}_0^n := \frac{1}{m} \sum_{i=1}^m
p_{0, i} \eqdef \bar{p}_0$. For the optimal case, we use the result from
\cref{thm:one-slow-sol}, in which we replace $s_1$ with $\bar{p}_0$, which
results in estimator $\hat{\prm}^*_0$.

\myparagraph{Bias and Shift}
\cref{fig: bias_shift} depicts the bias and shift of $\hat{\theta}^*_0$
with one standard deviation confidence intervals.
For $\alpha > 0$, the confidence intervals shrink very fast with
$m$ for big values of $p_0$ due to the shrinking introduced by $\clip$
function.

\myparagraph{Loss} Now, we analyze the loss of $\hat{\prm}^*_0$.

\begin{theorem}
    \label{theorem: expected_loss}
    The expected loss of $\hat{\theta}_0^\ast$ for $\alpha \le 0$ is
    \[
        \E_{z \sim D^*_1}((\hat{\theta}^*_0 - z)^2) = \frac{(1 -
        \abs{\alpha})^2}{1-2\alpha} \Par[\Big]{\frac{1 - 4(m+1)p_0}{4m}} +
        \frac{1}{4}.
    \]
    For all values of $\alpha$, the expected loss converges to the
    optimal expected loss: $\lim_{m \to \infty} \E((\hat{\theta}_0^\ast - z)^2) =
        \E(({\theta}_0^\ast - z)^2).$
\end{theorem}

\begin{figure}[ht]
    \includegraphics[width=\linewidth]{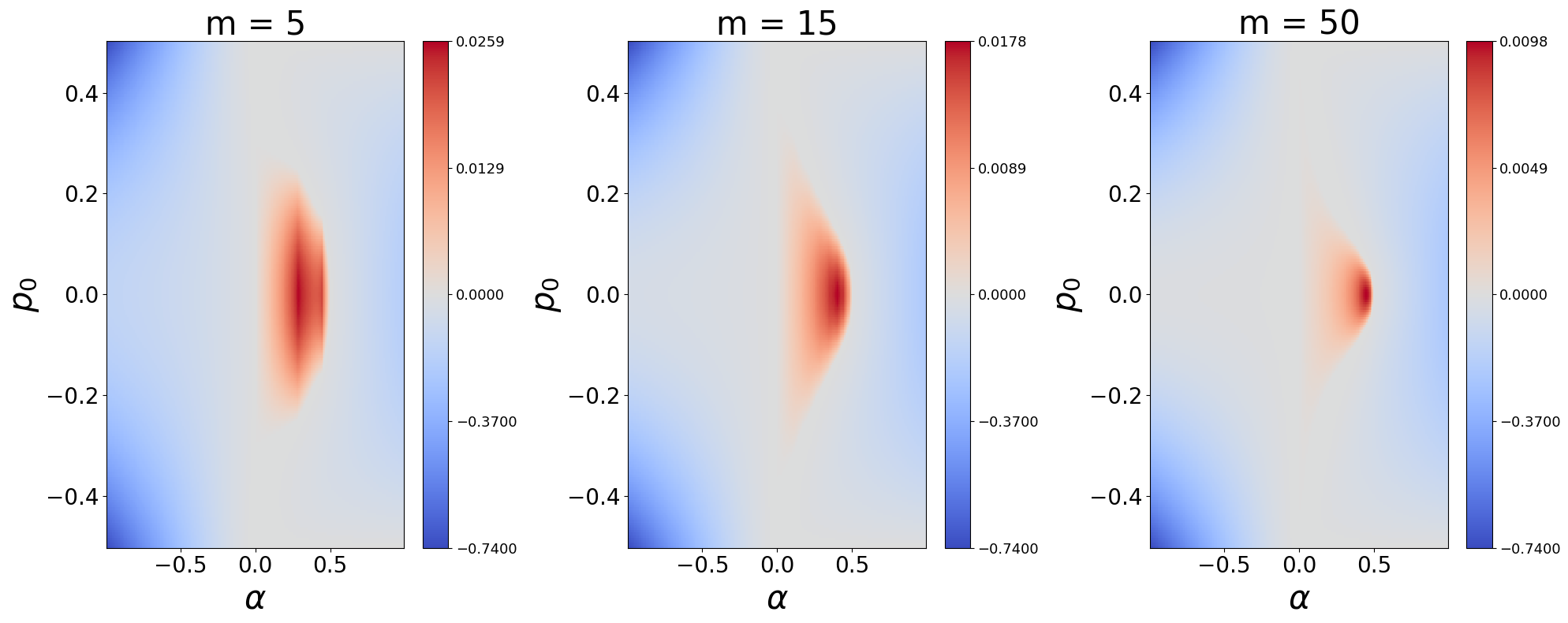}
    \caption{The dependence of the differences in expected
    losses, $\E(\loss(\hat{\prm}^*_0) - \loss(\hat{\prm}^n_0))$, on $p_0$ and
    $\alpha$, for different $m$.}
    \label{fig: loss}
\end{figure}

We discuss the loss for all values of $\alpha$ in \cref{sec:proofs}. To visualize the results of \cref{theorem: expected_loss}, we plot the
difference in expected losses between $\hat{\theta}_0^\ast$ and
$\hat{\theta}_0^n$ in \cref{fig: loss}. Due to random sampling, we observe a region where $\hat{\theta}_0^n$ outperforms
$\hat{\theta}_0^\ast$. However,
for larger values of $m$, this region diminishes.

\subsubsection{RL Simulations}
\label{sec:one-rl}

In this section, we check whether our results in the perfect information case
transfer to the general performative prediction problem with information
restrictions. In this setting, we consider episodic exploration and
additionally assume that $\lambda = 0$ is known to the provider. In this case,
the samples from the second period after deployment allow the model provider to
estimate the performativity parameters.

We implement Algorithm 1 of \citet{l22w} with hyperparameter
$\beta=2^{-8}$ to find the optimal predictions. We visualize the prediction path
of the algorithm in \cref{fig:rl} (left). After some exploration episodes, the
predictions of the model provider and the means of the distribution quickly
converge to the theoretically predicted values, which validates our results in the perfect information case.

\begin{figure}[ht]
    \input{fig4-rl.pgf}
    \caption{The predictions, $\prm_t$, (blue) the means, $p_t$, (orange)
    and their theoretical equilibrium values (red and green,
    respectively) in RL setting over episodes (left) or time (right) for $\pi =
    0.2$, $\alpha = 0.15$, $\gamma = 0.9$, and $m=100$, where $m$ is the number
    of samples observed from test distribution at each step. The left and right
    plots correspond to the $T=1$ slow episodic setting (with $\lambda = 0$)
    and $T=\infty$ slow setting (with $\lambda = 0.3$), respectively.}
    \label{fig:rl}
\end{figure}

\section{Infinite Horizon Model}
\label{sec:infinite_horizon}

\begin{figure*}[ht]
    \input{fig5-inf.pgf}
    \caption{The plots depict the dependence of $\prm^*_0$ (blue), $p^*_1$
    (orange), $s_1$ (green), and $p^*_\infty$ (red) on $p_0$ for $\lambda =
    0.8$, $\pi = 0.2$, and $\gamma=0.5$ in $T=\infty$ case. Columns correspond
    to the different $\alpha$, the top and bottom rows correspond to the slow
    and rapid cases, respectively.}
    \label{fig:inf-sol}
\end{figure*}

Now, we study the long-term effects of performativity by analyzing our model
for $T=\infty$. We first theoretically study the perfect information case and then use simulations to analyze the case of unknown problem parameters.

\subsection{Perfect Information}
\label{sec:inf-slow}
\subsubsection{Slow Deployment}

\begin{theorem}[Proof in \cref{sec:proof-inf-slow-sol}]
    \label{thm:inf-slow-sol}
    Assume that the PRM path does not take extreme values $\forall t \:
    \abs{\prm^*_t} \neq 1/2$ and $1 - 2 \alpha \ge \sqrt{\gamma} \beta$. Then,
    the solution to the problem (\ref{eq:opt-cont-prob}) in the $T=\infty$ slow deployment
    case satisfies
    \[
        \frac{\prm^*_t - \prm^*_\infty}{p_0 - p^*_\infty} = \frac{2 (1 -
        \abs{\alpha}) \lambda}{1 - 2 \alpha + \xi} \omega^t, \: \frac{p^*_t -
        p^*_\infty}{p_0 - p^*_\infty} = \omega^t, \text{ where }
    \]
    \[
        \begin{split}
            \prm^*_\infty & \defeq \frac{(1 - \gamma \beta) (1 - \abs{\alpha} -
            \beta) \pi}{1 - 2 \alpha - \beta + \alpha \beta - \gamma
            \beta (1 - \alpha - \beta)},\\
            p^*_\infty &\defeq \frac{(1 - \alpha - \gamma \beta) (1 -
            \abs{\alpha} - \beta) \pi}{1 - 2 \alpha - \beta + \alpha
            \beta - \gamma \beta (1 - \alpha - \beta)},\\
            \omega & \defeq \beta + \frac{2 \alpha \beta}{1 - 2 \alpha + \xi},
            \xi \defeq \sqrt{1 - \frac{4 \alpha (1 - \alpha)}{1 - \gamma
            \beta^2}}.
        \end{split}
    \]
\end{theorem}

Notice that the restriction $\forall t \: \abs{\prm^*_t} \neq 1/2$ could hold
only if $\omega \le 1$. There is an upper bound on $\alpha$ beyond which the
model provider is incentivized to choose the extreme values of $\prm_t$. So, if
this bound does not hold, after some time, the model provider always benefits
from setting $\abs{\prm^*_t} = 1/2$, even though this prediction is necessarily
biased. Additionally, if $\omega < 1$, the solution converges $\prm^*_t \to
\prm^*_\infty, s^*_t \to s^*_\infty, p^*_t \to p^*_\infty$ in the limit $t \to
\infty$, allowing us to study the long-term effects of PRM.

We visualize the solution for all cases in \cref{fig:inf-sol} (top row). The
restriction $\forall t \: \abs{\prm^*_t} \neq 1/2$ does not cover
the cases of big positive values of $\alpha$. In such scenarios, the PRM
prediction depends on $p_0$ discontinuously because the model provider has a
strong incentive to shift the mean to extreme values.

For the rest of this section, we assume that $\pi > 0$.

\myparagraph{Long-Term Bias} The long-term bias follows
\[
    \prm^*_\infty - p^*_\infty = \frac{\alpha (1 - \abs{\alpha} - \beta) \pi}{1
    - 2 \alpha - \beta + \alpha \beta - \gamma \beta (1 - \alpha - \beta)}.
\]
Even in the limit $t \to \infty$, the PRM solution has a non-vanishing
bias. If $\alpha > 0$ and $\alpha$ is small, the long-term bias is positive.
Even though the bias increases the error term in \cref{eq:mse}, the model
provider benefits in terms of uncertainty because the biased prediction shifts the
mean to more extreme values. On the other hand, if $\alpha < 0$ and
$\abs{\alpha}$ is small, the bias is negative. In the negative feedback case,
the negative bias again shifts the mean to more extreme values than the
unbiased prediction, reducing uncertainty.

\myparagraph{Long-Term Shift} The long-term shift of $\prm^*_t$ is non-zero:
\[
    p^*_\infty - \pi = \frac{(\alpha - \abs{\alpha} + \alpha \abs{\alpha} +
    \gamma \beta (\abs{\alpha} - \alpha)) \pi}{1 - 2 \alpha - \beta + \alpha
    \beta - \gamma \beta (1 - \alpha - \beta)}.
\]

\myparagraph{Comparison with Naive Path} We have that $\prm^n_\infty = p^n_\infty = \frac{1 - \abs{\alpha} - \beta}{1 - \alpha -
    \beta} \pi$. The bias of the naive path tends to zero as $t \to \infty$. The long-term shift
is also zero if $\alpha > 0$. If $\alpha < 0$,
\[
    \frac{p^*_\infty - \pi}{p^n_\infty - \pi} = \frac{1 - \gamma \beta +
    \frac{\abs{\alpha}}{2}}{1 - \gamma \beta + \frac{\abs{\alpha}}{1 +
    \abs{\alpha} / (1 - \beta)}} < 1.
\]
The long-term shift of the naive path is bigger than that of the PRM path
in the negative feedback case.

Similarly to $T=1$, the naive path has a smaller bias and shift than the PRM
path in the positive feedback case, while the PRM path has a smaller shift
in the negative feedback case. However, the long-term bias of the naive path is
$0$, even in the negative feedback case.

\subsubsection{Rapid Deployment}
\label{sec:inf-rapid}

\begin{theorem}[Proof in \cref{sec:proof-inf-rapid-sol}]
    \label{thm:inf-rapid-sol}
    Assume that the PRM path does not take extreme values $\forall
    t \: \abs{\prm^*_t} \neq 1/2$. Then, the solution to the problem
    (\ref{eq:opt-cont-prob}) in $T=\infty$ rapid case satisfies
    \[
        \prm^*_t = \frac{2}{1 + \chi} (p_0 - p^*_\infty) \kappa^t +
        \prm^*_\infty,\:
        p^*_t = (p_0 - p^*_\infty) \kappa^t + p^*_\infty,
    \]
    where $\kappa \defeq \beta + \frac{2 \alpha}{1 + \chi}$, $\chi \defeq \sqrt{1 - \frac{4 \gamma \alpha (\alpha + \beta)}{1 -
            \gamma \beta^2}}$ and
    \[
        \begin{split}
            \prm^*_\infty & \defeq \frac{(1 - \gamma \beta) (1 - \abs{\alpha} -
            \beta) \pi}{1 - \alpha - \beta - \gamma (\alpha + \beta -
            \beta (2 \alpha + \beta))},\\
            p^*_\infty & \defeq \frac{(1 - \gamma (\alpha + \beta)) (1 -
            \abs{\alpha} - \beta) \pi}{1 - \alpha - \beta - \gamma
            (\alpha + \beta - \beta (2 \alpha + \beta))}.
        \end{split}
    \]
\end{theorem}

Similarly to the slow case, the restriction $\forall t \: \abs{\prm^*_t} \neq
1/2$ could hold only if $\abs{\kappa} \le 1$. If $\abs{\kappa} < 1$,
$\prm^*_\infty$ and $p^*_\infty$ represent the long-term values of $\prm^*_t$
and $p^*_t$, respectively.

\cref{fig:inf-sol} (bottom row) visualizes the solution for all cases.
Again, the assumption $\forall t \: \abs{\prm^*_t} \neq 1/2$ does not cover
large $\abs{\alpha}$. If $\abs{\alpha}$ is large, the PRM
prediction depends discontinuously on $p_0$. If $\alpha > 0$, the mean,
depending on $p_0$, converges to one of two equilibrium values. If $\alpha <
0$, the mean oscillates between two values that correspond to extreme
predictions.

For the rest of this section, we assume that $\pi > 0$.

\myparagraph{Long-Term Bias} We get
\[
    \prm^*_\infty - p^*_\infty = \frac{\gamma \alpha (1 - \abs{\alpha} - \beta)
    \pi}{1 - \alpha - \beta - \gamma (\alpha + \beta - \beta (2 \alpha +
    \beta))}.
\]
The bias is again not zero and behaves similarly to the slow case for small
$\abs{\alpha}$.

\myparagraph{Long-Term Shift} We get a non-zero long-term shift:
\[
    p^*_\infty - \pi = \frac{(\alpha - \abs{\alpha} + \gamma (\alpha
    \abs{\alpha} + (\abs{\alpha} - \alpha) \beta) \pi}{1 - \alpha - \beta -
    \gamma (\alpha + \beta - \beta (2 \alpha + \beta))}.
\]

\myparagraph{Comparison with Naive Path} Notice that the mean in the naive path case satisfies $p^n_{t+1} = \alpha
p^n_{t-1} + \beta p^n_t + (1 - \abs{\alpha} - \beta) \pi$. Since $\alpha +
\beta < 1$, the mean converges to an equilibrium, which satisfies $\prm^n_\infty = p^n_\infty = \frac{1 - \abs{\alpha} - \beta}{1 - \alpha -
    \beta} \pi$.
Again, the long-term bias of the naive path is zero. The shift is zero if
$\alpha > 0$. If $\alpha < 0$,
\[
    \frac{p^*_\infty - \pi}{p^n_\infty - \pi}
    = \frac{1 + \gamma \abs{\alpha} / 2 - \gamma \beta}{1 + \gamma \abs{\alpha}
    / 2 - \gamma \beta + \frac{\gamma \abs{\alpha} (1 - \abs{\alpha} -
    \beta)}{2 (1 + \abs{\alpha} - \beta)}} < 1.
\]
Similarly to the slow case, the shift is smaller for $\prm^*_t$.

\subsection{RL Simulations}
\label{sec:inf-rl}

Finally, we check whether our results in the perfect information case
transfer to the general performative prediction problem with information
restrictions. We consider a usual sequential RL problem. We implement a simple heuristic algorithm, which learns the
performative response by deploying extreme predictions $\{-1/2, 1/2\}$ at
random for the first $4$ steps. Then the model provider learns the parameters of
the performative response by likelihood maximization and deploys the optimal
policy under the resulting estimates. We visualize the prediction path of the algorithm
in \cref{fig:rl} (right). After some exploration, the predictions and the means of the distribution quickly converge to the
theoretically-predicted equilibrium values, which validates our theoretical
analysis of the perfect information case.

\section{Discussion and Future Work}
\label{sec:conclusion}

Our results suggest that the performatively optimal (PRM) path is, in general,
biased and introduces a non-zero mean shift. These effects are more expressed
when the mean responds positively to model predictions or when it responds
negatively, but the model is updated rapidly and the performativity is high. To
understand the potential impact of such effects, we now provide two example
scenarios and interpret our measures and technical results in a social context.

\myparagraph{Case study: drug efficacy estimation}
Consider a scenario in which a company is trying to estimate the effectiveness
of a drug they produce against a specific disease. We define our binary random
variables as indicators that the drug cures a randomly sampled patient. To
model the well-known placebo effect, under which beliefs about the
effectiveness of a drug may further increase its positive impact, we assume a
positive performative response ($\alpha > 0$). Consider the one-period positive
feedback model in Section \ref{sec:one_period}. Then, a positive/negative bias
indicates an exaggerated/understated prediction of the average drug efficacy
respectively, which may make it harder to find the most effective drug on the
market. At the same time, a positive shift indicates a higher drug efficacy due
to the placebo effect, which is, of course, desirable for combating the
disease.

Our results in Section 4 with $p_0 = \pi$ suggest that whenever $p_0 > 0$ (i.e.
the drug is effective to begin with), PRM would lead to a positive bias, i.e.
exaggerated prediction on the drug's effectiveness; as well as positive shift
and thus increased drug effectiveness due to performativity.

\myparagraph{Case study: traffic prediction}
Consider a model provider seeking to predict which of the two roads, A or B, is
less busy. We model this by defining the binary random variable as an indicator
for the event that road A is less busy. Consider our infinite horizon negative
feedback model. Positive or negative bias of PRM corresponds to the model
provider redirecting more traffic to road A or B respectively. At the same
time, positive or negative shift indicates an increase in the usage of road B
or A respectively. The bias is probably an undesirable property of the
prediction as it makes some drivers choose a sub-optimal road. At the same
time, the shift might be benign or adverse, depending on the context.

In the slow deployment case, the mean usage of roads becomes more equalized
(Figure 5, top-right part) compared to the case when no performativity is
present, which is intuitively desirable. In the rapid deployment case, if the
strength of performativity is small, the usage becomes more equalized (Figure
5, bottom row, third plot). If the performativity is large, the usage
oscillates between roads (Figure 5, bottom row, fourth plot), which may be
undesirable.

\myparagraph{Limitations and future work}
In this work, we focus on mean estimation of binary variables only and work
under the linear response model (\ref{eqn:linear_response}). This makes the
analysis of the long-term dynamics driven by (\ref{eq:opt-cont-prob}) tractable
and allows for defining natural metrics of impact and interpreting them in
context. Despite its simplicity, we
hope that our model can be qualitatively useful in broader settings. First, the
linear response naturally arises as a first-order Taylor approximation for any
performative response. Thus, our results may (at least qualitatively) transfer
to situations of weak performative response. Second, as noted in the discussion
of Lemma 3.2, the error-uncertainty decomposition holds for a broad class of
distributions. Thus, we can expect PRM to generally prefer distributions with
smaller aleatoric uncertainty. For example, in the case of multinomial
distribution, the model provider has an additional incentive to concentrate the probability mass
on a small subset of outcomes.

Additionally, our results can easily be extended to the following more general
group setting. Imagine that clients consist of several independent groups, and
each group reacts to the predictions of the model in the same way as the whole
distribution in our paper. Also, assume that the model provider additionally
observes covariates that are predictive for group membership before making a
prediction. This modification makes our problem much closer to the usual
supervised learning tasks where the model provider needs to simultaneously
learn a model for membership prediction and outcomes for each group. At the
same time, our results in the perfect information setting can be directly
transfered to this setup by independently applying the previous analysis to each
group. The main limitation of such an extension is the assumption that groups
evolve independently. This assumption could hold in the setting of drug
efficacy prediction, but it will probably not hold in traffic prediction.

We hope that our work will encourage further analysis of the broader impact of
PRM . In particular, it would be interesting to analyze more complex
distributions (e.g., in a regression setting) and models of performative
response.

\section*{Acknowledgments}
This research was partially funded from the Ministry of Education and Science of Bulgaria (support for INSAIT, part of the Bulgarian National Roadmap for Research Infrastructure). The authors thank Kristian Minchev for his helpful feedback and discussions on this work.

\section*{Impact Statement}

This paper presents work whose goal is to advance the field of Machine Learning. Our theoretical analysis contributes to better understanding of the impact of machine learning on society. Therefore, we expect that our results can serve a positive purpose in increasing the awareness about ML impact and encouraging further research on related topics. There are many potential societal consequences of our work, none which we feel must be specifically highlighted here.

\bibliography{ta-bib}

\newpage
\onecolumn

\appendix

\begin{center}
    {\LARGE Supplementary Material}
\end{center}

\begin{itemize}
    \item \cref{sec:add-res} contains additional results.
    \item \cref{sec:proofs} contains proofs of all results.
    \item \cref{sec:add-rl} contains details of RL experiments.
\end{itemize}

\section{Additional Results}
\label{sec:add-res}

This section presents additional results that were not included in the main
text.

\subsection{Generalization of Impact Metrics}
\label{sec:gen-metrics}

This section discusses possible generalizations of the impact metrics defined
in \cref{subsec:measures_of_impact}

\myparagraph{Mean Shift Metric}
Regarding the mean shift metric, we identify two possible extensions, for
parametric and non-parametric distributions, respectively.

Consider a setting where the distribution is parametrized by a finite number of
parameters, $D_t = D(\mathbf{w}_t)$, where $\mathbf{w}_t = (w^1_t, \dots,
w^k_t)$. This is, for example, the case for distributions defined via causal
graphical models, as well as for many common distributions (e.g., exponential
families). Then, one can define the parameter shift metric as $\mathtt{shift}_t
= \E(\mathbf{w}_t - \mathbf{w}^0_t)$, where $\mathbf{w}^0_t$ are the parameters
of the distribution at time $t$ if the distribution was not affected by the
performativity.

In a non-parametric setting, we can instead define a divergence-based metric
$\mathtt{shift}_t = \E(K(D_t, D^0_t))$, where $K$ is an arbitrary divergence
function (e.g., KL-divergence). The function $K$ can be designed to capture an
undesirable shift in the distribution, according to the target application.

These metrics can then be studied under different models for the performative
response and counterfactual dynamics of the distribution in the absence of
performativity, which can be chosen depending on the learning task and
application under consideration.

\myparagraph{Bias Metric}
Regarding the bias metric, we provide two extensions suitable for cases where
the distribution is divided into several groups, which is relevant, e.g., in
fairness-sensitive applications.

First, in a setting with several groups and multi-labeled data, one could
calculate a matrix of biases with one entry for each group and label
defined as follows $\mathtt{bias}^{g,y} = \E(\E_{(X, Y, G) \sim
D^{test}_t}(q^y_t(X) - [Y=y] | G=g))$, where $q_t(X) = (q^1_t(X), \dots,
q^{|\mathcal{Y}|}_t(X))$ is the vector of model's softmax probabilities at time
$t$ and $G$ is the group. These biases could be interpreted as a measure of
unfairness among groups.

Second, one can use established metrics from the literature of bias
amplification see \citet{z17m,w21d,z23m,t24m}.

\subsection{Perfect Information}

This section contain additional results in the perfect information setting.

\subsubsection{One-Period Slow Deployment, Perfect Information}

Here, we present additional results for \cref{sec:one-slow}.

\paragraph{Comparison with Naive Path in Symmetric Case}

In the symmetric case, we assume that the equilibrium probability is
symmetric, $\pi = 0$. Then,
\[
    \begin{aligned}
        \loss^n_0 &= 1/4 + (1 - 2 \alpha - 2 \beta) p_0^2,
        & \loss^*_0 &= 1/4 - \frac{\beta^2}{1 - 2 \alpha} p_0^2,\\
        \bias^n_0 &= (1 - \alpha - \beta) p_0,
        & \bias^*_0 &= \frac{\alpha \beta}{1 - 2 \alpha} p_0,\\
        \shift^n_1 &= (\alpha - \abs{\alpha} \lambda) p_0,
        & \shift^*_1 &= \frac{(\alpha - \abs{\alpha} + \alpha \abs{\alpha})
        \lambda}{1 - 2 \alpha} p_0.
    \end{aligned}
\]
If $\alpha > 0$, we get
\[
    \frac{\abs{\bias^*_0}}{\abs{\bias^n_0}} = \frac{\alpha \lambda}{(1
    - 2 \alpha) (1 - \lambda)}.
\]
If the performativity, $\alpha$, or the inertia, $\lambda$, is big then the
naive prediction is preferable in terms of bias. Otherwise, the optimal
prediction is preferable. (The same analysis holds for the shift of
estimator.)

If $\alpha \le 0$, we get
\[
    \frac{\abs{\bias^*_0}}{\abs{\bias^n_0}} = \frac{\abs{\alpha} (1 -
    \abs{\alpha}) \lambda}{(1 + 2 \abs{\alpha}) (1 - \lambda + \abs{\alpha} +
    \abs{\alpha} \lambda)} < 1.
\]
Thus, the optimal prediction is preferable in terms of bias. (The same is true
for the shift.)

Finally, if $1 - 2 \alpha > 0$, the loss penalty equals to
\[
    \Par*{1 - 2 \alpha - 2 \beta + \frac{\beta^2}{1 - 2 \alpha}} p_0^2.
\]
To analyze it consider two cases: $\lambda = 0$ and $\lambda = 1$. If $\lambda
= 0$, we get that the following penalty
\[
    \loss^n_0 - \loss^*_0 = (1 - 2 \alpha) p_0^2.
\]
This penalty is bigger than the penalty in the equilibrium case for small
$\alpha$. If $\lambda = 1$, we get that $p_0 = s_1$. So, we get the same answer
as in the equilibrium case.

\subsubsection{Two-Period Slow Deployment, Perfect Information}

Here, extend \cref{sec:one-slow} by solving the two-period case and comparing
with it.

\begin{proposition}[Proof in \cref{sec:proof-two-slow-sol}]
    \label{thm:two-slow-sol}
    Assume that $1 - 2 \alpha > \sqrt{\gamma} \abs{\alpha} \beta$. Then, the
    solution to the problem (\ref{eq:opt-cont-prob}) in $T=2$ slow case
    satisfies
    \[
        \prm^*_0 = \clip\Par*{\frac{(1  -  \abs{\alpha}) ((1  -  2
        \alpha  +  \gamma \alpha \beta^2) s_1 + \gamma \alpha \beta (1 -
        \lambda) \pi)}{(1 - 2 \alpha)^2 - \gamma \alpha^2 \beta^2},
        -\frac{1}{2}, \frac{1}{2}},
    \]
    if $2 (1 - \abs{\alpha}) \abs{s^*_2} \le 1 - 2 \alpha$ (which always holds
    for $\alpha \le 0$).
\end{proposition}

We visualize whole solution on \cref{fig:fin-sols}, bottom row. Notice that on
the left part of the picture we operate in regime $1 - 2 \alpha < \sqrt{\gamma}
\abs{\alpha} \beta$. In this situation, the optimal prediction depends
non-continuously on $p_0$ because of the incentive to push the mean to the
extremes. Additionally notice that the left plot has a kink on its right side.
This kink corresponds to the transition between the cases $2 (1 - \abs{\alpha})
\abs{s^*_2} \le 1 - 2 \alpha$ and $2 (1 - \abs{\alpha}) \abs{s^*_2} > 1 - 2
\alpha$.

If $\abs{\prm^*_0} < 1/2$ in the setting of \cref{thm:two-slow-sol}, we get
\[
    p^*_1 = \frac{(1 - \abs{\alpha}) ((1 - 2 \alpha) (1 - \alpha) s_1 +
    \gamma \alpha^2 \beta (1 - \lambda) \pi)}{(1 - 2 \alpha)^2 - \gamma
    \alpha^2 \beta^2}.
\]
For the rest of the subsubsection we assume $\prm^*_0 < 1/2$.

\paragraph{Bias of $\prm^*_0$}

We get
\[
    \prm^*_0 - p^*_1 = \frac{\alpha (1 - \abs{\alpha}) ((1 - 2 \alpha + \gamma
    \beta^2) s_1 + \gamma (1 - \alpha) \beta (1 - \lambda) \pi)}{(1 - 2
    \alpha)^2 - \gamma \alpha^2 \beta^2}.
\]
For equilibrium and symmetric $\pi$, the bias of prediction becomes more
pronounced because
\[
    \frac{1 - 2 \alpha + \gamma \beta^2}{(1 - 2 \alpha)^2 - \gamma \lambda^2
    \beta^2} \ge \frac{1}{1 - 2 \alpha}.
\]

\paragraph{Shift of $\prm^*_0$}

We get
\[
    p^*_1 - s_1 = \frac{((1 - 2 \alpha) (\alpha - \abs{\alpha} + \alpha
    \abs{\alpha}) + \gamma \alpha^2 \beta^2) s_1 + \gamma \alpha^2 (1 -
    \abs{\alpha}) \beta (1 - \lambda) \pi}{(1 - 2 \alpha)^2 - \gamma \alpha^2
    \beta^2}.
\]

\paragraph{Discussion}

We can see that generally the bias of the optimal prediction is exacerbated in
the two-period model. It happens because the motivation of the model provider
to skew the distribution becomes stronger due to longer horizon.

\paragraph{Comparison with Naive Path}

In equilibrium case $\pi = p_0$, given that the impact and bias of the naive
path is the same as for the one-period model and our results above, we get that
the naive path is even more preferable to the optimal path if $\alpha > 0$ in
terms of bias and shift. For the case of $\alpha < 0$, we get
\[
    \begin{split}
        -\frac{\bias^*_0}{\abs{\alpha} s_1} &= -\frac{(1 \! - \!
        \abs{\alpha}) (1 + 2 \abs{\alpha} + \gamma \beta (1 + \abs{\alpha} - 2
        \abs{\alpha} \lambda))}{(1 + 2 \abs{\alpha})^2 - \gamma \abs{\alpha}^2
        \beta^2},\\
        \frac{\shift^*_1}{\abs{\alpha} s_1} &= \frac{-(1 + 2 \abs{\alpha}) (2
        + \abs{\alpha}) + \gamma \abs{\alpha} (1 - \abs{\alpha}) \beta}{(1 + 2
        \abs{\alpha})^2 - \gamma \abs{\alpha}^2 \beta^2},\\
        \frac{\shift^n_1}{\abs{\alpha} s_1} &= -\frac{\bias^n_0}{\abs{\alpha}
        s_1} = -2.
    \end{split}
\]
By direct calculation, the bias and shift of the naive path is always higher
than those of the optimal path.

In the symmetric case $\pi = 0$, if $\alpha > 0$, we get
\[
    \frac{\abs{\bias^*_0}}{\abs{\bias^n_0}} = \frac{\alpha \lambda (1 + \gamma
    \beta^2 / (1 - 2 \alpha))}{((1 - 2 \alpha) - \gamma \alpha^2 \beta^2 / (1 -
    2 \alpha)) (1 - \lambda)}
\]
Notice that ratio $\beta^2 / (1 - 2 \alpha) = \lambda^2 (1 + \alpha^2 / (1 - 2
\alpha))$ is increasing in $\alpha$. Thus, the ratio of biases is increasing in
$\alpha$ and $\lambda$. So, similarly, to the one-period case, the optimal path
is preferable to the naive path in terms of bias if $\alpha$ and $\lambda$ are
small enough. (Same analysis holds for the shift.)

If $\alpha < 0$, we get
\[
    \frac{\abs{\shift^*_1}}{\abs{\shift^n_1}} = \frac{\abs{\alpha} \lambda (1
    + 2 \abs{\alpha} + \gamma \beta^2)}{((1 + 2 \abs{\alpha})^2 - \gamma
    \alpha^2 \beta^2) (1 + \abs{\alpha} - \lambda + \abs{\alpha} \lambda)}.
\]
This ratio is increasing in $\lambda$ and $\gamma$. Hence,
\[
    \frac{\abs{\shift^*_1}}{\abs{\shift^n_1}} \le \frac{2 + \alpha^2}{2 + 8
    \abs{\alpha} + 3 \alpha^2} \le 1.
\]
So, the bias of the optimal path is smaller than the bias of the naive path.
Similarly, the shift of the optimal path is smaller than the impact of the
naive path.

\paragraph{Discussion}

Similarly to the one-period case, the naive path might be preferable in terms
of bias and impact to the optimal path for $\alpha \ge 0$. However, for $\alpha
< 0$, the optimal path is superior to the naive path in terms of bias and
shift.

\subsubsection{Infinite Horizon Slow Deployment, Perfect Information}

This section contains additional results for \cref{sec:inf-slow}.

\paragraph{Bias of $\prm^*_0$}

We get
\[
    \bias^*_0 = \frac{(1 - \xi) (1 - \abs{\alpha})}{1 - 2 \alpha +
    \xi} s_1.
\]
Notice that, if $1 - 2 \alpha \ge \sqrt{\gamma} \beta$, then this bias is
bigger than in the two-period case. Thus, as previously, the longer time
horizon incentivizes the model provider to give more biased predictions.

\paragraph{Shift of $\prm^*_0$}

We get
\[
    \shift^*_1 = \frac{2 \alpha - \abs{\alpha} - \abs{\alpha} \xi}{1 - 2
    \alpha + \xi} s_1.
\]
Similarly to the bias of $\prm^*_0$, the impact of $\prm^*_0$ increases
compared to the two-period case if $\alpha > 0$. However, if $\alpha < 0$, the
impact becomes smaller than the two-period impact.

\paragraph{Bias and Shift of $\prm^n_0$}

The bias and impact of the naive path in the symmetric case follows
\[
    \begin{split}
        \bias^n_0 &= (1 - \alpha - \beta) p_0,\\
        \shift^n_1 &= (\alpha - \abs{\alpha} \lambda) p_0.
    \end{split}
\]

The bias of the naive path is smaller if
\[
    2 \alpha + \beta (1 + \gamma (\alpha + \beta) (1 - \alpha - \beta)) \ge 1,
\]
which happens only if $\alpha > 0$ and $\alpha$ is sufficiently big. (The same
inequality holds for the shift.)

\subsubsection{Infinite Horizon Rapid Deployment, Perfect Information}

This section contains additional results for \cref{sec:inf-rapid}.

\paragraph{Bias of $\prm^*_0$}

We get
\[
    \prm^*_0 - p_0 = \frac{1 - \chi}{1 + \chi} p_0.
\]
Assuming that $p_0 > 0$, we get the following classification of the model
provider actions. In the case of $\alpha > 0$, we get that $\prm^*_0 > p_0$. If
$\alpha < 0$ and $\alpha + \beta > 0$, $\prm^*_0 < p_0$. Finally, if $\alpha +
\beta < 0$, $\prm^*_0 > p_0$ again.

\paragraph{Shift of $\prm^*_0$}

We get
\[
    p^*_1 - s_1 = \kappa p_0 - \lambda p_0 = \Par*{-\abs{\alpha} \lambda +
    \frac{2 \alpha}{1 + \chi}} p_0.
\]
Since $\kappa$ increases in $\alpha$ and $\kappa|_{\alpha=0} = \lambda$,
the shift increases in $\abs{\alpha}$.

\subsubsection{Additional Visualizations}

We visualize the solutions for $T=1$ rapid case, $T=2$ rapid case, and $T=2$
slow case in \cref{fig:fin-sols}. As we can see, if $\alpha > 0$, the
prediction and the resulting next-period mean shift to more extreme values.
Otherwise, the prediction and mean shift to $0$ (the effect is more pronounced
for the mean).

\begin{figure}[ht]
    \input{fig6-fin-sols.pgf}
    \caption{The plots depict the dependence of $\prm^*_0$ (blue), $p^*_1$
    (orange), and $s_1$ (green) on $p_0$ for $\lambda = 0.8$, $\pi = 0.2$, and
    $\gamma=0.5$. Columns correspond to the different values of $\alpha$; the
    top row corresponds to the $T=1$ rapid case; the middle row corresponds to
    $T=2$ rapid case; the bottom row corresponds to the $T=2$ slow case.}
    \label{fig:fin-sols}
\end{figure}

\subsection{Imperfect Information, \texorpdfstring{$T=1$}{T=1} Slow Deployment}

Here, we present additional results for \cref{sec:imperf-one-slow}.

\subsubsection{Bias and Mean Shift Theoretical Results}

The bias for the naive estimator $\hat{\theta}_0^n$ is given by
\begin{equation*}
    \bias_0^n = p_0(|\alpha| - \alpha).
\end{equation*}
For the performative estimator $\hat{\theta}_0^\ast$, the bias is 
\begin{equation*}
    \bias_0^\ast= (1 - \alpha) \mathbb{E}[\hat{\theta}_0^\ast] - (1 - |\alpha|) p_0.
\end{equation*}

For the naive estimator $\hat{\theta}_0^n$ the mean shift is  
\begin{equation*}
    \shift_1^n = p_0(\alpha - |\alpha|) = - \bias_0^n,
\end{equation*}
and for the performative estimator $\hat{\theta}_0^\ast$, we have
\begin{equation*}
    \shift_1^\ast = \alpha \E[\hat{\theta}_0^\ast] - |\alpha|p_0.
\end{equation*}

\subsubsection{General Version of Theorem \ref{theorem: expected_loss}}

Here, we present a result that generalizes \cref{theorem: expected_loss},
offering theoretical insights for all possible values of $\alpha \in (-1,1)$.
\begin{theorem}
    \label{theorem: expected_loss_full}
    For the naive estimator $\hat{\theta}_0^n$ the expected loss is 
    \begin{equation*}
        \E_{z \sim D_1^{test}}[(\hat{\theta}_0^n - z)^2] = p_0^2 (2 |\alpha| -
        2\alpha - 1) + (2\alpha - 1)\frac{4p_0^2 - 1}{4m} + \frac14,
    \end{equation*}
    and for the performative estimator $\hat{\theta}_0^\ast$, we have 
    \begin{align*}
        \E&[(\hat{\theta}_0^\ast - z)^2] =\\
        &\begin{cases}
            \frac{(1 - |\alpha|)^2}{1-2\alpha} \bigg( \frac{\frac14 - p_0^2}{m}
            - p_0^2 \bigg) + \frac14 & \alpha \in (-1, 0]\\
            p_0 (1 - |\alpha|) \big(2F_{m, p_0 + \frac12}(\frac{m}{2}) - 1\big)
            + \frac{1-\alpha}{2} & \alpha \in [0.5, 1)\\
            \sum_{x \in I}((1 - 2\alpha)g(x)^2 - 2(1 - |\alpha|)p_0 g(x)) p(x) +
            (p_0(1-|\alpha|) - \frac{1-2\alpha}{4})F_{m, p_0 +
            \frac{1}{2}}\bigl(\frac{2 - 3\alpha}{2 - 2\alpha}m \bigr) & \\
            + (p_0(1-|\alpha|) + \frac{1-2\alpha}{4})F_{m, p_0 +
            \frac{1}{2}}\bigl( \frac{\alpha m}{2 - 2\alpha} \bigr) - p_0
            (1-|\alpha|) + \frac{1-\alpha}{2}, & \alpha \in (0, 0.5),
        \end{cases}
    \end{align*}
    where $I$ is the set of integers in $\big(\frac{\alpha m}{2-2\alpha},
    \frac{(2-3\alpha)m}{2-2\alpha} \big]$, $g(x) \defeq (\frac{1-\alpha}{1-2\alpha})(\frac{x}{m} -
    \frac{1}{2})$, $F_{m, p_0 + \frac12} (x) :=
    \sum_{k=0}^{\lfloor x \rfloor} p(x),$ and
    \begin{equation*}
        p(x) := \binom{m}{x} \bigg(\frac12 + p_0\bigg)^x \bigg(\frac12 -
        p_0\bigg)^{m-x}
    \end{equation*}

    Asymptotically, we have that as $m \to \infty$
    \begin{equation*}
        \E[(\hat{\theta}_0^\ast - z)^2] \to \verb|loss|_0^\ast
    \end{equation*}
    i.e. as $m$ goes to infinity, $\hat{\theta}_0^\ast$ approaches the optimal
    estimator for the risk minimisation problem.
\end{theorem}

\clearpage

\section{Proofs}
\label{sec:proofs}

\subsection{Proof of Lemma \ref{lemma: bias-variance}}
\begin{proof}
    Using conditional expectation we have
    \begin{align*}
        \E[(\theta_t - z)^2 \mid \theta_t, p_t^{test}] &= \E[\theta_t^2 - 2
        \theta_t  z + z^2 \mid  \theta_t, p_t^{test}]\\
        &= \theta_t^2 - 2 \theta_t \E[z \mid \theta_t, p_t^{test}] + \E[z^2
        \mid \theta_t, p_t^{test}]\\
        &= \theta_t^2 - 2 \theta_t p_t^{test} + \frac14\\
        &= (\theta_t^2 - p_t^{test})^2 + \frac14 - (p_t^{test})^2
    \end{align*}
\end{proof}

\subsection{Proof of Proposition \ref{thm:one-slow-sol}}
\label{sec:proof-one-slow-sol}

By direct calculation,
\[
    \prm_0^2 - 2 \prm_0 (\alpha \prm_0 + (1 - \abs{\alpha}) s_1) = (1 - 2 \alpha) \prm_0^2
    - 2 (1 - \abs{\alpha}) s_1 \prm_0.
\]

If the parabola above opens downwards, $1 - 2 \alpha \le 0$, it achieves
minimum at the extreme point of the domain. By analyzing both extreme points,
we get $\prm^*_0 = \sign(s_1) / 2$.

If the parabola opens upwards, $1 - 2 \alpha > 0$, it achieves the minimum at
the point in our domain closest to the vertex point. Thus, $\prm^*_0 =
\clip\Par*{\frac{(1 - \abs{\alpha}) s_1}{1 - 2 \alpha}, -\frac{1}{2},
\frac{1}{2}}$.

\subsection{Proof of Proposition \ref{thm:two-rapid-sol}}
\label{sec:proof-two-rapid-sol}

Going backward, we get
\[
    \prm^*_1 = p_1,
\]
which results in the following problem
\[
    \min_{\prm_0, \prm_1, p_1} \prm_0^2 - 2 \prm_0 p_0 - \gamma p_1^2 \:
    \text{s.t.} \: p_1 = \alpha \prm_0 + (1 - \abs{\alpha}) (\lambda p_0 + (1 -
    \lambda) \pi), \prm_t \in [\nicefrac{-1}{2}, \nicefrac{1}{2}].
\]
Similarly to \cref{sec:proof-one-slow-sol}, we get
\[
    \prm^*_0 = \clip\Par*{\frac{(1 + \gamma \alpha \beta)
    p_0 + \gamma \alpha (1 - \abs{\alpha}) (1 - \lambda) \pi}{1 - \gamma
    \alpha^2}, -\frac{1}{2}, \frac{1}{2}}.
\]
(Notice that $1 - \gamma \alpha^2 > 0$, which reduces the number of cases.)

\subsection{Proof of Theorems \ref{theorem: expected_loss} and \ref{theorem:
expected_loss_full}}

Before computing the expected loss, we first show the following result
regarding the first two moments of the performative estimator.

\begin{lemma}[Moments of the Performative Estimator] \label{lemma: moments}
    For the performative estimator $\hat{\theta}_0^\ast$, we have that the
    first two moments are given by
    \begin{equation*}
        \E[\hat{\theta}_0^\ast] = 
        \begin{cases}
            \frac{(1 - |\alpha|) p_0}{1-2\alpha} & \alpha \in (-1, 0]\\
            \frac12 - F_{m, p_0 + \frac12} (\frac{m}{2}) & \alpha \in [0.5,
            1)\\
            \sum_{x \in I} \big(\frac{1 - \alpha}{1-2\alpha}\big)
            \big(\frac{x}{m} - \frac12 \big) p(x) & \\
            + \frac12  - \frac12 F_{m, p_0 + \frac12}\big( \frac{2 -
            3\alpha}{2-2\alpha}m \big) & \\
            - \frac12 F_{m, p_0 + \frac12}\big( \frac{\alpha }{2-2\alpha}m\big)
            & \alpha \in (0, 0.5)
        \end{cases}
    \end{equation*}
    and 
    \begin{equation*}
        \E[(\hat{\theta}_0^\ast)^2] = 
        \begin{cases}
            \big( \frac{1-|\alpha|}{1-2\alpha}\big)^2 \big( \frac{0.25 -
            p_0^2}{m} + p_0^2 \big) & \alpha \in (-1, 0]\\
            \frac14 & \alpha \in [0.5, 1)\\
            \sum_{x \in I} \big(\frac{1 - \alpha}{1-2\alpha}\big)^2
            \big(\frac{x}{m} - \frac12 \big)^2 p(x)  & \\
            + \frac14  - \frac14 F_{m, p_0 + \frac12}\big( \frac{2 -
            3\alpha}{2-2\alpha}m \big) & \\
            + \frac14 F_{m, p_0 + \frac12}\big( \frac{\alpha }{2-2\alpha}m\big)
            & \alpha \in (0, 0.5)
        \end{cases}
    \end{equation*}
    where $I$ is the set of integers in $\big(\frac{\alpha m}{2-2\alpha},
    \frac{(2-3\alpha)m}{2-2\alpha} \big]$, $F_{m, p_0 + \frac12} (x) :=
    \sum_{k=0}^{\lfloor x \rfloor} p(x),$ and
    \begin{equation*}
        p(x) := \binom{m}{x} \bigg(\frac12 + p_0\bigg)^x \bigg(\frac12 -
        p_0\bigg)^{m-x}
    \end{equation*}
\end{lemma}
\begin{proof}[Proof of Lemma \ref{lemma: moments}]
    Recall that $\hat{\theta}_0^\ast$ is given by
    \[
        \prm^*_0 =
        \begin{cases}
            \clip\Par[\big]{\frac{(1 - \abs{\alpha}) \bar{p}_0}{1 - 2
            \alpha},
            -\frac{1}{2}, \frac{1}{2}}, & 1 - 2 \alpha > 0,\\
            \sign(\bar{p}_0) / 2, & 1 - 2 \alpha \le 0.
        \end{cases}
    \]
    We consider three cases for the value of $\alpha$: 
    \begin{enumerate}[(i)]
        \item $\alpha \in (-1, 0]$

            In this case we have 
            \begin{equation*}
                \prm^\ast_0 = \frac{1 - |\alpha|}{1-2\alpha} \bar{p}_0
            \end{equation*}
            and therefore 
            \begin{equation*}
                \E[\prm^\ast_0] = \frac{1 - |\alpha|}{1-2\alpha}
                \bar{p}_0, \quad \E[(\prm^\ast_0)^2] = \bigg( \frac{1 -
                |\alpha|}{1-2\alpha}\bigg)^2\E[\bar{p}_0^2] = \bigg(
                \frac{1 - |\alpha|}{1-2\alpha}\bigg)^2\bigg( p_0^2 +
                \frac{\frac14 - p_0^2}{m}\bigg),
            \end{equation*}
            where we have used that $p_{0, i} \sim D_0$ for $i=1, \dots, m$,
            and thus $p_{0,i} + \frac12$ follows a Bernoulli distribution with
            parameter $p_0 + \frac12$.

        \item $\alpha \in [0.5, 1)$

            In this case, we have that 
            \begin{equation*}
                \prm^\ast_0 = 
                \begin{cases}
                    \frac12 & \bar{p}_0 \ge 0\\
                    -\frac12 & \bar{p}_0 <  0.
                \end{cases}
            \end{equation*}
            Since $\bar{p}_0 = \bar{q} - \frac12$, where
            $\bar{q} := \frac{1}{m}\sum_{i=1}^m q_i$ and $q_i := p_{0,
            i}$, so that $q_i \sim Bern(p_0 + \frac12)$, we know that the
            events can be written as
            \begin{align*}
                \{ \bar{p}_0 \ge 0 \} = \{ \bar{q} \ge 0.5 \}, \quad
                \{ \bar{p}_0 < 0 \} = \{ \bar{q} < 0.5 \}.
            \end{align*}
            Therefore, 
            \begin{align*}
                \prm^\ast_0 = \frac12 \chi_{\{ \bar{q} \ge 0.5 \}} -
                \frac12 \chi_{\{ \bar{q} < 0.5 \}}.
            \end{align*}
            Finally, using the law of total expectation, we get that
            \begin{align*}
                \E[\prm^\ast_0] &= \E[\prm^\ast_0 | \bar{q} \ge 0.5]
                \Pr[\bar{q} \ge 0.5] + \E[\prm^\ast_0 |\bar{q}
                < 0.5] \Pr[\bar{q} < 0.5]\\
                &= \frac12 \Pr[\bar{q} \ge 0.5] - \frac12
                \Pr[\bar{q} < 0.5]\\
                &= \frac12 - F_{m, p_0 + \frac12}(0.5m),
            \end{align*}
            where we have used that $m \bar{q} \sim Bin(m ,p_0 + 0.5)$.
            Similarly for the second moment
            \begin{align*}
                \E[(\prm^\ast_0)^2] &= \E[(\prm^\ast_0)^2 | \bar{q} \ge
                0.5] \Pr[\bar{q} \ge 0.5] + \E[(\prm^\ast_0)^2
                |\bar{q} < 0.5] \Pr[\bar{q} < 0.5]\\
                &= \frac14 \Pr[\bar{q} \ge 0.5] + \frac14 \Pr[\bar{q}
                < 0.5]\\
                &= \frac14.
            \end{align*}
        \item $\alpha \in (0, 0.5)$

            In this case we have 
            \begin{align*}
                \prm^\ast_0 &= 
                \begin{cases}
                    \frac{1 - \alpha}{1-2\alpha} \bar{p}_0, & \text{if }
                    \bar{p}_0 \in \big(- \frac{1-2\alpha}{2 - 2\alpha},
                    \frac{1-2\alpha}{2 - 2\alpha} \big] \eqdef A\\
                    \frac12 , & \text{if } \bar{p}_0 > \frac{1-2\alpha}{2
                    - 2\alpha} \eqdef B\\
                    -\frac12 , & \text{if } \bar{p}_0 \le -
                    \frac{1-2\alpha}{2 - 2\alpha} \eqdef C
                \end{cases}
            \end{align*}
            where we have denoted by $A,B,C$ the random events that we have not
            clipped the value of the performative estimator, that we have
            clipped it from above or that we have clipped in from below. Using
            the law of total expectation, we have
            \begin{align*}
                \E[\prm^\ast_0] &= \E[\prm^\ast_0 | A] \Pr[A] + \E[\prm^\ast_0
                | B] \Pr[B] + \E[\prm^\ast_0 | C] \Pr[C]\\
                &= \E[\prm^\ast_0 \chi_{A}] + \frac12 \Pr[B] - \frac12 \Pr[C]\\
                &= \E[\prm^\ast_0 \chi_{A}] + \frac12 \Pr\bigg[\bar{q} >
                \frac{2-3\alpha}{2-2\alpha}\bigg] - \frac12
                \Pr\bigg[\bar{q} \le \frac{\alpha}{2-2\alpha}\bigg]
            \end{align*}
            The first term can be computed as follows
            \begin{align*}
                \E[\prm^\ast_0 \chi_{A}] &= \sum_{x \in I} \frac{1 -
                \alpha}{1-2\alpha} \bigg( \frac{x}{m} - \frac12 \bigg) p(x),
            \end{align*}
            where we have used that $m\bar{p}_0 + m/2 \sim Bin(m, p_0 +
            \frac12)$ and have denoted by $p(x)$ the PMF of $Bin(m, p_0 +
            \frac12$. The last two terms are easily expressed via the CDF of
            the same distribution, giving us that
            \begin{align*}
                \E[\prm^\ast_0] = \sum_{x \in I} \bigg(\frac{1 -
                \alpha}{1-2\alpha}\bigg) \bigg(\frac{x}{m} - \frac12 \bigg)
                p(x)
                + \frac12  - \frac12 F_{m, p_0 + \frac12}\bigg( \frac{2 -
                3\alpha}{2-2\alpha}m \bigg)
                - \frac12 F_{m, p_0 + \frac12}\bigg( \frac{\alpha
                }{2-2\alpha}m\bigg).
            \end{align*}
            where $I$ is the set of integers in the interval $( \frac{\alpha
            }{2 - 2\alpha}m, \frac{2-3\alpha}{2 - 2\alpha} m]$. Similarly, for
            the second moment we have that
            \begin{align*}
                \E[(\prm^\ast_0)^2] &= \E[(\prm^\ast_0)^2 | A] \Pr[A] +
                \E[(\prm^\ast_0)^2 | B] \Pr[B] + \E[(\prm^\ast_0)^2 | C]
                \Pr[C]\\
                &= \E[(\prm^\ast_0)^2 \chi_{A}] + \frac14 \Pr[B] + \frac14
                \Pr[C]\\
                &= \E[(\prm^\ast_0)^2 \chi_{A}] + \frac14 \Pr\bigg[\bar{q}
                > \frac{2-3\alpha}{2-2\alpha}\bigg] - \frac12
                \Pr\bigg[\bar{q} \le \frac{\alpha}{2-2\alpha}\bigg]\\
                &= \sum_{x \in I} \bigg(\frac{1 - \alpha}{1-2\alpha}\bigg)^2
                \bigg(\frac{x}{m} - \frac12 \bigg)^2 p(x)
                + \frac14  - \frac14 F_{m, p_0 + \frac12}\bigg( \frac{2 -
                3\alpha}{2-2\alpha}m \bigg)
                + \frac14 F_{m, p_0 + \frac12}\bigg( \frac{\alpha
                }{2-2\alpha}m\bigg),
            \end{align*}
            which finishes the proof.
    \end{enumerate}
\end{proof}

Now, we are ready to present the full proof.

\begin{proof}[Proof of Theorems \ref{theorem: expected_loss} and \ref{theorem:
    expected_loss_full}]
    We begin by rewriting the expected loss as follows 
    \begin{align*}
        \E[(\theta_0 - z_0)^2] &= \E[\E[\theta_0^2 - 2\theta_0 z_0 + z_0^2 |
        \theta_0]]\\
        &= \E[\theta_0^2 - 2 \theta_0 \E[z_0 | \theta_0] + \E[z_0^2 |
        \theta_0]]\\
        &= \E\bigg[\theta_0^2 - 2 \theta_0 p_1(\theta_0) + \frac14\bigg]\\
        &=(1-2\alpha)\E[\theta_0^2] - 2 (1 - |\alpha|)p_0 \E[\theta_0] +
        \frac14
    \end{align*}
    where the expectation is only in terms of the randomness of the
    observations $\{p_{0, i} \}_{i=1}^m$.

    For the naive estimator, $\hat{\theta}_0^n$, we have that the first two
    moments are
    \begin{align*}
        \E[\hat{\theta}_0^n] & = p_0\\
        \E[(\hat{\theta}_0^n)^2] &= p_0^2 + \frac{(\frac12 - p_0)(\frac12 +
        p_0)}{m},
    \end{align*} 
    which follows since $p_{0,i} \sim D_0$ for $i = 1, \dots, m$. Therefore, we
    get
    \begin{align*}
        \E[(\hat{\theta}_0^n - z)^2] &=  
        (1-2\alpha)\E[\theta_0^2] - 2 (1 - |\alpha|)p_0 \E[\theta_0] +
        \frac14\\
        &= p_0^2 (2 |\alpha| - 2\alpha - 1) + \frac14 + \frac{(2\alpha - 1)(4
        p_0^2 - 1)}{4m}
    \end{align*}

    For the performative estimator, we use the first and second moments of
    $\hat{\theta}_0^\ast$ from \cref{lemma: moments} to obtain
    \begin{align*}
        \E&[(\hat{\theta}_0^\ast - z)^2] =\\
        &\begin{cases}
            \frac{(1 - |\alpha|)^2}{1-2\alpha} \bigg( \frac{\frac14 - p_0^2}{m}
            - p_0^2 \bigg) + \frac14 & \alpha \in (-1, 0]\\
            p_0 (1 - |\alpha|) \big(2F_{m, p_0 + \frac12}(\frac{m}{2}) - 1\big)
            + \frac{1-\alpha}{2} & \alpha \in [0.5, 1)\\
            \sum_{x \in I}((1 - 2\alpha)g(x)^2 - 2(1 - |\alpha|)p_0 g(x)) p(x) +
            (p_0(1-|\alpha|) - \frac{1-2\alpha}{4})F_{m, p_0 +
            \frac{1}{2}}\bigl(\frac{2 - 3\alpha}{2 - 2\alpha}m \bigr) & \\
            + (p_0(1-|\alpha|) + \frac{1-2\alpha}{4})F_{m, p_0 +
            \frac{1}{2}}\bigl( \frac{\alpha m}{2 - 2\alpha} \bigr) - p_0
            (1-|\alpha|) + \frac{1-\alpha}{2}, & \alpha \in (0, 0.5),
        \end{cases}
    \end{align*}
    where $g(x) \defeq (\frac{1-\alpha}{1-2\alpha})(\frac{x}{m} -
    \frac{1}{2})$.

    Asymptotically, as $m \to \infty$, we have that the moments of
    $\hat{\theta}_0^\ast$ for $\alpha \in (-1, 0]$ are given by
    \begin{align*}
        \E[\hat{\theta}_0^\ast] &= \frac{(1-|\alpha|)}{1-2\alpha} p_0 \to
        \frac{(1-|\alpha|)}{1-2\alpha} p_0\\
        \E[(\hat{\theta}_0^\ast)^2] &= \frac{(1-|\alpha|)^2}{(1-2\alpha)^2}
        \bigg( \frac{0.25 - p_0^2}{m} +  p_0^2\bigg) \to
        \frac{(1-|\alpha|)^2}{(1-2\alpha)^2} p_0^2
    \end{align*}
    Similarly, for $\alpha \in [0,5, 1)$, we have 
    \begin{align*}
        \E[\hat{\theta}_0^\ast] &= \frac12 - F_{m, p_0 +
        \frac12}\bigg(\frac{m}{2}\bigg) \to \frac{sign(p_0)}{2}\\
        \E[(\hat{\theta}_0^\ast)^2] &= \frac14 \to \frac14
    \end{align*}
    where we have used that the CDF function $F_{m, p_0 +
    \frac12}\bigg(\frac{m}{2}\bigg)$ converges to $1$ for non-negative $p_0$
    and to $0$ for negative $p_0$ as $m\to \infty$.

    Finally, for $\alpha \in (0, 0.5)$, we have that 
    \begin{align*}
        \E[\hat{\theta}_0^\ast] &= \E\bigg[\clip{\bigg( \frac{1 - |\alpha|
        p_0}{1-2\alpha}, -\frac12, \frac12 \bigg)}\bigg]\\
        &= \E\bigg[ \frac{(1-|\alpha|)\bar{p}_0}{1-2\alpha}
        \chi_{\{\bar{p}_0 \in A\}} \bigg] + \frac12 \Pr[\bar{p}_0 \in
        B  ] - \frac{1}{2} \Pr[\bar{p}_0 \in C ]\\
        &\to \frac{(1-|\alpha|){p_0}}{1-2\alpha}  \chi_{\{{p_0} \in A\}} +
        \frac12 \chi_{[{p_0} \in B  ]} - \frac{1}{2} \chi_{[{p_0} \in C ]}\\
        &= \E[{\theta}_0^\ast \mid \alpha \in (0, 0.5)].
    \end{align*}
    where $A$ denotes the region (a function of $\alpha$), where
    $\hat{\theta}_0^\ast$ has not been clipped, $B$ represents the region where
    it has been clipped from above, and $C$ is the region where it has been
    clipped from below. The third line follows from: (1) the law of large
    numbers, which ensures that $\bar{p}_0 \to p_0$ almost surely as $m
    \to \infty$, and (2) the dominated convergence theorem. The same argument
    applies for $\E[(\hat{\theta}_0^\ast)^2]$. Thus, combining this with the
    other two cases for $\alpha$, we get the following asymptotic results
    \begin{align*}
        \lim_{m \to \infty} \E[\hat{\theta}_0^\ast] = \theta_0^\ast, \quad
        \lim_{m \to \infty} \E[(\hat{\theta}_0^\ast)^2] = (\theta_0^\ast)^2.
    \end{align*}
    Therefore, we can conclude that as $m\to \infty$,
    \begin{equation*}
        \E[(\hat{\theta}_2^\ast - z)^2] \to \verb|loss|_0^\ast.
    \end{equation*}
\end{proof}

\subsection{Proof of Theorem \ref{thm:inf-slow-sol}}
\label{sec:proof-inf-slow-sol}

Consider Lagrangian function
\begin{multline*}
    L(\vec{w}, \vec{q}, \vec{\nu}, \vec{\mu}, \vec{\eta}) \defeq\\
    \sum_{t=0}^\infty \gamma^t (\prm_t^2 - 2 \prm_t p_{t+1}) - (\alpha \prm_t + \beta p_t
    + (1 - \abs{\alpha}) (1 - \lambda) \pi - p_{t+1}) \nu_t - (1 / 2 - \prm_t)
    \mu_t - (\prm_t + 1 / 2) \eta_t.
\end{multline*}
KKT conditions for this infinite-horizon problem \citep[see Section 4.5
of][]{s89r} give
\[
    \begin{aligned}
        0 &= 2 \gamma^t (\prm_t - p_{t+1}) - \alpha \nu_t + \mu_t - \eta_t,\\
        0 &= -2 \gamma^t \prm_t + \nu_t - \beta \nu_{t+1},\\
        0 &= (1 / 2 - \prm_t) \mu_t, \mu_t \ge 0,\\
        0 &= (\prm_t + 1 / 2) \eta_t, \eta_t \ge 0.
    \end{aligned}
\]
Thus, the solution for the case when the restrictions on $\prm_t$ are non-binding
satisfies
\[
    \begin{aligned}
        \prm_{t+1} &= \frac{(1 - 2 \alpha + \gamma \alpha \beta^2)}{\gamma (1 -
        \alpha) \beta} \prm_t - \frac{1 - \gamma \beta^2}{\gamma (1 - \alpha)
        \lambda} s_{t+1} + \frac{\beta (1 - \lambda)}{(1 - \alpha) \lambda}
        \pi,\\
        s_{t+2} &= \alpha \lambda \prm_t + \beta s_{t+1} + (1 - \lambda) \pi.
    \end{aligned}
\]

We get that the optimal path satisfies a first-order linear recurrence relation
for $\prm_t$ and $s_t$. Its characteristic equation follows
\[
    x^2 - \frac{1 - 2 \alpha + \gamma \beta^2}{\gamma (1 - \alpha) \beta} x +
    \frac{1}{\gamma} = 0.
\]
It gives the following eigenvalues
\[
    x_{0,1} = \frac{1 - 2 \alpha + \gamma \beta^2 \pm
    \sqrt{(1 - \gamma \beta^2) ((1 - 2 \alpha)^2 - \gamma \beta^2)}}{2 \gamma
    (1 - \alpha) \beta}.
\]
Notice that the product of these eigenvalues is $1/\gamma$. Thus, one of the
eigenvalues is necessarily bigger than $1$ in absolute value. Due to the
restrictions on $w$, the homogeneous solution corresponding to this eigenvalue
should be zero.

Consider the case of $1 - 2 \alpha \ge \sqrt{\gamma} \beta$, then, the smallest
eigenvalue, $\omega$, satisfies
\[
    \omega = \beta + \frac{(1 - 2 \alpha) (1 - \gamma \beta^2) - \sqrt{(1 -
    \gamma \beta^2) ((1 - 2 \alpha)^2 - \gamma \beta^2)}}{2 \gamma (1 - \alpha)
    \beta} = \beta + \frac{2 \alpha \beta}{1 - 2 \alpha + \xi}.
\]
Corresponding eigenvector gives the following homogeneous solution
\[
    s^h_{t+1} = s \omega^t, \:
    \prm^h_t = \frac{2 (1 - \abs{\alpha})}{1 - 2 \alpha + \xi} r \omega^t.
\]
One of inhomogeneous solutions satisfies
\[
    s^i_{t+1} = \frac{(1 - 2 \alpha - \gamma \beta (1 - \alpha)) (1 - \lambda)
    \pi}{1 - 2 \alpha - \beta + \alpha \beta - \gamma \beta (1 - \alpha -
    \beta)}, \:
    \prm^i_t = \frac{(1 - \gamma \beta) (1 - \abs{\alpha}) (1 - \lambda) \pi}{1
    - 2 \alpha - \beta + \alpha \beta - \gamma \beta (1 - \alpha - \beta)}.
\]
Using the initial conditions, we get the desired solution.

\subsection{Proof of Theorem \ref{thm:inf-rapid-sol}}
\label{sec:proof-inf-rapid-sol}

Consider Lagrangian function
\begin{multline*}
    L(\vec{w}, \vec{q}, \vec{\nu}, \vec{\mu}, \vec{\eta}) \defeq\\
    \sum_{t=0}^\infty \gamma^t (\prm_t^2 - 2 \prm_t p_t) - (\alpha \prm_t + \beta p_t
    + (1 - \abs{\alpha}) (1 - \lambda) \pi - p_{t+1}) \nu_t - (1 / 2 - \prm_t)
    \mu_t - (\prm_t + 1 / 2) \eta_t.
\end{multline*}
KKT conditions for this infinite-horizon problem \citep[see Section 4.5
of][]{s89r} give
\[
    \begin{aligned}
        0 &= 2 \gamma^t (\prm_t - p_t) - \alpha \nu_t + \mu_t - \eta_t,\\
        0 &= -2 \gamma^t \prm_t + \nu_{t-1} - \beta \nu_t,\\
        0 &= (1 / 2 - \prm_t) \mu_t, \mu_t \ge 0,\\
        0 &= (\prm_t + 1 / 2) \eta_t, \eta_t \ge 0.
    \end{aligned}
\]
Thus, the solution for the case when the restrictions on $\prm_t$ are non-binding
satisfies
\[
    \begin{aligned}
        \prm_{t+1} &= \frac{1 + \gamma \alpha \beta}{\gamma (\alpha + \beta)} \prm_t
        - \frac{1 - \gamma \beta^2}{\gamma (\alpha + \beta)} p_t + \frac{\beta
        (1 - \abs{\alpha}) (1 - \lambda)}{\alpha + \beta} \pi,\\
        p_{t+1} &= \alpha \prm_t + \beta p_t + (1 - \abs{\alpha}) (1 - \lambda)
        \pi.
    \end{aligned}
\]

We get that the optimal path satisfies a first-order linear recurrence relation
for $\prm_t$ and $p_t$. Its characteristic equation follows
\[
    x^2 - \frac{1 + \gamma \beta (2 \alpha + \beta)}{\gamma (\alpha + \beta)} x
    + \frac{1}{\gamma} = 0.
\]
It gives the following eigenvalues
\[
    x_{0,1} = \frac{1 + \gamma \beta (2 \alpha + \beta) \pm
    \sqrt{(1 - \gamma \beta^2) (1 - \gamma (2 \alpha + \beta)^2)}}{2 \gamma
    (\alpha + \beta)}.
\]
Notice that the product of these eigenvalues is $1/\gamma$. Thus, one of the
eigenvalues is necessarily bigger than $1$ in absolute value. Due to the
restrictions on $w$, the homogeneous solution corresponding to this eigenvalue
should be zero.

The smallest eigenvalue, $\kappa$, satisfies
\[
    \kappa = \frac{1 + \gamma \beta (2 \alpha + \beta) - \sqrt{(1 - \gamma
    \beta^2) (1 - \gamma (2 \alpha + \beta)^2)}}{2 \gamma (\alpha + \beta)} =
    \beta + \frac{2 \alpha}{1 + \chi}.
\]
Thus, homogeneous solution follows
\[
    q^h_t = q \kappa^t, \: \prm^h_t = \frac{2}{1 + \chi} q \kappa^t.
\]
One of inhomogeneous solutions satisfies
\[
    q^i_t = \frac{(1 - \gamma (\alpha + \beta)) (1 - \abs{\alpha}) (1 -
    \lambda) \pi}{1 - \alpha - \beta - \gamma (\alpha + \beta - \beta (2
    \alpha + \beta))},\:
    \prm^i_t = \frac{(1 - \gamma \beta) (1 - \abs{\alpha}) (1 - \lambda)
    \pi}{1 - \alpha - \beta - \gamma (\alpha + \beta - \beta (2 \alpha +
    \beta))}.
\]

\subsection{Proof of Proposition \ref{thm:two-slow-sol}}
\label{sec:proof-two-slow-sol}

Using the results of \cref{thm:one-slow-sol}, we get
\[
    \prm^*_1 =
    \begin{cases}
        \clip\Par*{\frac{(1 - \abs{\alpha}) s_2}{1 - 2 \alpha}, -\frac{1}{2},
        \frac{1}{2}}, & 1 - 2 \alpha > 0,\\
        \frac{\sign(s_2)}{2}, & 1 - 2 \alpha \le 0,
    \end{cases}
\]
which results in the following loss in the second period:
\[
    (\prm^*_1)^2 - 2 \prm^*_1 p^*_2 =
    \begin{cases}
        -\frac{(1 - \abs{\alpha})^2 s_2^2}{1 - 2 \alpha},
        & 2 (1 - \abs{\alpha}) \abs{s_2} < 1 - 2 \alpha,\\
        \frac{1 - 2 \alpha}{4} - (1 - \abs{\alpha}) \abs{s_2},
        & 2 (1 - \abs{\alpha}) \abs{s_2} \ge 1 - 2 \alpha.
    \end{cases}
\]
Notice that
\[
    (\prm^*_1)^2 - 2 \prm^*_1 p^*_2 \ge - \frac{(1 - \abs{\alpha})^2 s_2^2}{1 -
    2 \alpha}.
\]
Thus,
\[
    \sum_{t=0}^1 \gamma^t (\prm_t^2 - 2 \prm_t p_{t+1}) \ge \prm_0^2 - 2
    \prm_0 p_1 - \frac{\gamma (1 - \abs{\alpha})^2 s_2^2}{1 - 2 \alpha}.
\]
So, if the minimizer of the right hand side satisfies $2 (1 - \abs{\alpha})
\abs{s^{\text{rhs}, *}_2} \le 1 - 2 \alpha$, it will minimize the left-hand
side.

Similarly to \cref{sec:proof-one-slow-sol}, we have that the minimizer of the
right-hand side satisfies
\[
    \begin{split}
        & \prm^{\text{rhs}, *}_0 =\\
        &
        \begin{cases}
            \clip\Par[\big]{\frac{(1 - \abs{\alpha}) ((1 - 2 \alpha + \gamma
            \alpha \beta^2) s_1 + \gamma \alpha \beta (1 - \lambda) \pi)}{(1
            - 2 \alpha)^2 - \gamma \alpha^2 \beta^2}, -\frac{1}{2},
            \frac{1}{2}},
            & 1 - 2 \alpha > \sqrt{\gamma} \abs{\alpha} \beta,\\
            \frac{\sign((1 - 2 \alpha + \gamma \alpha \beta^2) s_1 + \gamma
            \alpha \beta (1 - \lambda) \pi)}{2},
            & 1 - 2 \alpha \le \sqrt{\gamma} \abs{\alpha} \beta.
        \end{cases}
    \end{split}
\]

Thus, when $1 - 2 \alpha > \sqrt{\gamma} \abs{\alpha} \beta$ and $2 (1 -
\abs{\alpha}) \abs{s^*_2} \le 1 - 2 \alpha$, we get
the desired solution to our problem.

\newpage

\section{Details of RL Simulations}
\label{sec:add-rl}

This section gives additional details about RL-like simulations in Sections
\ref{sec:one-rl} and \ref{sec:inf-rl}.

\subsection{One-period Episodic Simulations}

We consider episodic exploration of $T=1$ slow model, where we additionally
assume that $\lambda = 0$ and $\lambda$ is known to the provider. In this
setting, we assume that each episode has the following structure.
\begin{enumerate}
    \item Nature samples $q_0$.
    \item The provider observes $\{z^i_0\}_{i=0}^{m-1} \sim D_0^m$.
    \item The provider deploys $\prm_0$.
    \item The provider observes $\{z^i_1\}_{i=0}^{m-1} \sim D_1^m$.
\end{enumerate}

We implement \cref{alg:rl-episodic}, adopted version of Algorithm 1 of
\citet{l22w}, where we denote the episode number by $\tau$, with hyperparameter
$\beta=2^{-8}$ to find the optimal prediction. (Notice that the first period
observations are non-informative for the log-likelihood maximization because
$\lambda = 0$.)

\begin{algorithm}[ht]
    \caption{Optimistic Maximum Likelihood Estimation}
    \label{alg:rl-episodic}
    \begin{algorithmic}
        \STATE {\bfseries Initialize:} $B^0 = \{(\alpha, \pi) : \alpha \in [-1,
        1], \pi \in [-1/2, 1/2]\}$, $\mathcal{D} = \{\}$
        \FOR{$\tau=0$ {\bfseries to} $T$}
            \STATE Deploy $\prm^\tau_0 = \argmin_{\prm \in [-\nicefrac{1}{2},
            \nicefrac{1}{2}]} \min_{(\alpha, \pi) \in
            B^\tau} \loss\Par*{\prm \given \alpha, \pi}$
            \STATE Observe $S^\tau_1 \sim (D^\tau_1)^m$
            \STATE Add $(\prm^\tau_0, S^\tau_1)$ to $\mathcal{D}$
            \STATE Update $
                B^{\tau+1} = \Bc[\bigg]{(\alpha, \pi) \in B^0 : \sum_{(\prm, S)
                \in D} \log \Pr\Par*{S \given \alpha, \pi, \prm} \ge
                \max_{(\alpha, \pi) \in B^0} \sum_{(\prm, S) \in D} \log
                \Pr\Par*{S \given \alpha, \pi, \prm} - \beta}
                $
        \ENDFOR
    \end{algorithmic}
\end{algorithm}

\subsection{Infinite Horizon Simulations}

We consider episodic exploration of $T=\infty$ slow model, where the provider
know the value of $\gamma$. In this setting, we assume that each step has the
following structure.
\begin{enumerate}
    \item The provider observes $\{z^i_t\}_{i=0}^{m-1} \sim D_t^m$.
    \item The provider deploys $\prm_t$.
\end{enumerate}
For this case, we implement heuristic \cref{alg:rl-heuristic}, where we denoted
the value function as
\[
    V(p_0, \alpha, \pi, \lambda, \gamma) \defeq
    \min_{(\prm_t)_{t=0}^\infty} \sum_{t=0}^\infty \gamma^t \loss\Par*{\prm_t
    \given \prm_{t-1}, \dots, \prm_0, p_0, \alpha, \pi, \lambda}.
\]
This algorithm learns the performative response by deploying extreme
predictions $\{-1/2, 1/2\}$ at random for the first $4$ steps. Then the model
provider learns the parameters of the performative response by likelihood
maximization and deploys the optimal policy under their estimates.
\begin{algorithm}[ht]
    \caption{Greedy Exploration}
    \label{alg:rl-heuristic}
    \begin{algorithmic}
        \STATE Observe $S_0 \sim D_0^m$
        \FOR{$t=0$ {\bfseries to} $3$}
            \STATE Deploy $\prm_t$ at random from $\{-1/2, 1/2\}$
            \STATE Observe $S_{t+1} \sim D_{t+1}^m$
        \ENDFOR
        \FOR{$t=4$ {\bfseries to} $T$}
            \STATE Estimate $(\alpha, \pi, \lambda, p_0) = \argmax_{\alpha,
            \pi, \lambda, p_0} \sum_{\tau=0}^t \log \Pr\Par*{S_t \given
            \prm_{t-1}, \dots, \prm_0, p_0, \alpha, \pi, \lambda}$
            \STATE Deploy $\prm_t = \argmin_{\prm_t}
            \loss\Par*{\prm_t \given p_{t+1}(\prm_t, \dots, \prm_0, p_0,
            \alpha, \pi, \lambda)} + \gamma V(p_{t+1}(\prm_t, \dots, \prm_0,
            p_0, \alpha, \pi, \lambda), \alpha, \pi, \lambda, \gamma)$
            \STATE Observe $S_{t+1} \sim D_{t+1}^m$
        \ENDFOR
    \end{algorithmic}
\end{algorithm}

\clearpage

\end{document}